%% file: main.tex
\documentclass[10pt]{article}

\usepackage[preprint]{tmlr}

\input{math_commands.tex}

\usepackage{hyperref}
\usepackage{url}

\usepackage[ruled,vlined]{algorithm2e}

\usepackage{wrapfig}
\usepackage{makecell}
\usepackage{verbatim}
\usepackage{booktabs,multirow}
\usepackage{siunitx}
\usepackage[table]{xcolor}
\usepackage{pifont}
\usepackage{enumitem}
\usepackage[most]{tcolorbox}
\usepackage{soul}
\usepackage{textgreek}
\usepackage{arydshln}
\usepackage{float}
\usepackage{placeins}

\usepackage{cleveref}
\usepackage{caption}
\usepackage{subcaption}
\usepackage{nicematrix}

\newcommand{\ourmodel}[1]{PANORAMA}
\newcommand{\ourdataset}[1]{PanoCaps}

\definecolor{headbg}{HTML}{F2F2F2}
\definecolor{groupblue}{HTML}{DAE8FC}
\definecolor{rowgray}{HTML}{F2F2F2}
\newlength{\groupinset}

\newcommand{\finding}[2]{
    \begin{tcolorbox}[
        colback=headbg,
        colframe=headbg,
        arc=5pt,
        boxsep=5pt,
        left=10pt,
        right=10pt,
        top=2pt,
        bottom=2pt,
        boxrule=0.8pt,
        drop shadow=gray!50!white,
        enhanced jigsaw
    ]
    \vspace{-0.15cm}
         #2
    \end{tcolorbox}
    \vspace{-0.15cm}
}

\title{PANORAMA: Panoptic Grounded Captioning via Mask \mbox{Proposal} Selection}

\author{\name Sara Pieri$^{1}$, Evangelos Kazakos$^{2}$, Shizhe Chen$^{1}$, Josef Sivic$^{2}$, Cordelia Schmid$^{1}$ \\
        \addr $^{1}$Inria, \'Ecole normale sup\'erieure, CNRS, PSL Research University \\
        $^{2}$Czech Institute of Informatics, Robotics and Cybernetics, Czech Technical University in Prague}

\begin{document}

\maketitle

\input{sec/00_abstract}

\input{sec/01_introduction}

\input{sec/02_related_work}
\input{sec/05_dataset}
\input{sec/03_pre}
\input{sec/04_method}
\input{sec/06_abla}
\input{sec/07_expr}

\FloatBarrier
\input{sec/08_conclusion}

\bibliographystyle{tmlr}
\bibliography{references}

\input{sec/09_appendix}

\end{document}

%% file: math_commands.tex
\usepackage{amsmath,amsfonts,bm}

\def\eqref#1{equation~\ref{#1}}

\def\1{\bm{1}}

\DeclareMathAlphabet{\mathsfit}{\encodingdefault}{\sfdefault}{m}{sl}
\SetMathAlphabet{\mathsfit}{bold}{\encodingdefault}{\sfdefault}{bx}{n}

%% file: sec/00_abstract.tex
\begin{abstract} 
Intelligent systems that act in the world require image understanding that is both comprehensive and spatially grounded.
Current vision-language models (VLMs) can generate fluent and detailed image captions, but reliably associating them with image pixels remains challenging. Existing methods that combine dense captioning with pixel-level grounding often produce either incomplete descriptions or inaccurate segmentation masks.
We study this problem through panoptic grounded captioning, a task that requires a VLM to describe both foreground objects and background regions while grounding each referring phrase with pixel-level masks. We make three contributions.
First, we introduce \textbf{\ourdataset{}}, a human-annotated benchmark constructed from panoptic segmentation datasets. It provides dense captions with near-complete pixel coverage and image-text alignments at the entity level, supporting both training and evaluation. We further propose a phrase-mask matching protocol and a generalized Panoptic Quality (gPQ) metric that jointly evaluates textual and mask agreement.
Second, we formulate phrase grounding as selection from a phrase-conditioned pool of mask proposals and introduce \textbf{\ourmodel{}}, a VLM that conditions a pretrained segmenter on contextualized phrase representations to obtain candidate masks and learns to select those corresponding to each phrase. Training this interface jointly with caption generation enables \ourmodel{} to produce high-quality masks while allowing each phrase to refer to a single region or multiple instances.
Third, \ourmodel{} achieves the best overall grounding on \ourdataset{} and matches or exceeds specialized models across several pixel-level grounding tasks.
Experiments show that our method produces precise entity-level segmentations while maintaining detailed, mask-consistent captions.
Code, data and models are available at \url{https://www.di.ens.fr/willow/research/panorama/}.
\end{abstract}

%% file: sec/01_introduction.tex
\section{Introduction}
\label{sec:intro}

Systems that perceive and act in the physical world require scene understanding that is both comprehensive and spatially precise. Such understanding involves not only describing what is present, but also spatially grounding the described entities within the scene. This capability is critical for embodied agents, robotic manipulation, visual assistance, and human-AI interaction~\citep{driess2023palme,brohan2023rt2,gurari2020captioningblind}, where incomplete or inaccurately localized descriptions can lead to unreliable downstream decisions.
Modern vision-language models (VLMs)~\citep{chen2024internvl2.5,bai2025qwen3,comanici2025gemini} have made substantial progress in visual understanding and generation. They can identify salient objects, describe complex scenes, and answer detailed questions in fluent natural language. However, reliably grounding these descriptions in the corresponding image regions remains challenging.

A natural approach is to combine dense image description with pixel-level grounding. The recently proposed \textbf{panoptic grounded captioning} task~\citep{deng2025coconutpancap} requires models to generate full-scene descriptions that cover both foreground objects and background regions, while grounding every referring phrase with a pixel-level mask. However, as shown in~\Cref{fig:teaser}, existing models fall short of this goal: they omit relevant entities, produce incomplete descriptions, or ground referring phrases with coarse and imprecise masks.
Progress on this task is limited by two key challenges. First, existing grounded captioning datasets~\citep{rasheed2024glamm,deng2025coconutpancap} typically trade off annotation coverage against quality. Automatically constructed datasets can provide dense supervision, but often suffer from noisy alignments, incomplete mask coverage, and inconsistent captions. In contrast, human-annotated datasets typically provide only short or partial descriptions, leaving substantial portions of the scene unlabeled. Consequently, existing benchmarks do not simultaneously provide the comprehensive and fine-grained supervision required to train and evaluate models for both dense image description and precise pixel-level grounding.
Second, directly predicting segmentation masks with a VLM places a substantial spatial-prediction burden on an architecture designed primarily for autoregressive language generation. Although large pretrained segmenters provide strong localization capabilities, effectively aligning their mask predictions with open-ended textual phrases remains challenging~\citep{rasheed2024glamm,yuan2025sa2va}.

\input{figures/teaser}

Motivated by these limitations, we introduce \textbf{\ourdataset{}}, a new benchmark comprising 3.5K human-annotated images for training and evaluating panoptic grounded captioning models. Built from diverse segmentation datasets~\citep{deng2024coconut,Zhou_2017_CVPR,Miao_2022_CVPR}, \ourdataset{} provides fine-grained phrase-mask alignments between human-written captions and image regions. As illustrated in~\Cref{fig:dataset}, it combines free-form, full-scene descriptions with near-complete mask coverage, addressing the fundamental trade-off between descriptive completeness and annotation quality in existing benchmarks.

We further propose \textbf{\ourmodel{}}, short for \textbf{PANO}ptic g\textbf{R}ounded c\textbf{A}ptioning via \textbf{MA}sk proposal selection. As illustrated in~\Cref{fig:method}, \ourmodel{} formulates phrase grounding as selection from a phrase-conditioned pool of mask proposals.
For each referring phrase, the VLM emits a \texttt{[SEG]} token whose hidden state is projected into a concept vector that conditions a pretrained segmenter~\citep{carion2025sam}. The segmenter generates candidate masks, and a learned scorer selects the proposals corresponding to the phrase. 
This design separates referent understanding from boundary prediction: the VLM resolves the intended entity using the full image-language context, the segmenter provides mask proposals, and a learned match scorer links the two by selecting the proposals corresponding to each phrase. In contrast, \texttt{[SEG]}-decoding approaches generate each mask directly from the token embedding, requiring a single representation to capture both the intended referent and its spatial extent.
Because the scorer selects a subset of proposals, our approach directly supports grounding phrases to a single region, multiple regions, or no segmentable region.

\ourmodel{} achieves the best overall grounding on \ourdataset{} and matches or exceeds specialized grounding models on grounded conversation generation~\citep{rasheed2024glamm}, referring expression segmentation~\citep{yu2016modeling}, and generalized grounding~\citep{liu2023gres,hu2025groundingsuite}. Finetuning existing models on \ourdataset{} consistently improves their grounding performance, demonstrating the value of dense, high-quality supervision. Beyond training, \ourdataset{} provides a new benchmark for evaluating comprehensive, pixel-level scene understanding.

\noindent Our contributions are:

\begin{itemize}
\item We introduce \textbf{\ourdataset{}}, a human-annotated panoptic grounded captioning benchmark with approximately $99\%$ pixel coverage, fine-grained phrase-mask alignments, free-form full-scene captions, and diverse image sources. We further provide an evaluation protocol and a generalized Panoptic Quality (gPQ) metric that extends PQ to free-form phrases and graded match quality.

\item We propose \textbf{\ourmodel{}}, which formulates phrase grounding as selection from phrase-conditioned proposals, decoupling semantic identification from boundary delineation to enable accurate and flexible pixel-level grounding.

\item We achieve strong results on panoptic grounded captioning on \ourdataset{} and across several related grounding tasks, demonstrating that well-localized pixel-level predictions can be achieved without sacrificing detailed, mask-consistent language generation.
\end{itemize}

%% file: figures/teaser.tex
\begin{figure}[t]
    \centering
    \includegraphics[width=\linewidth]{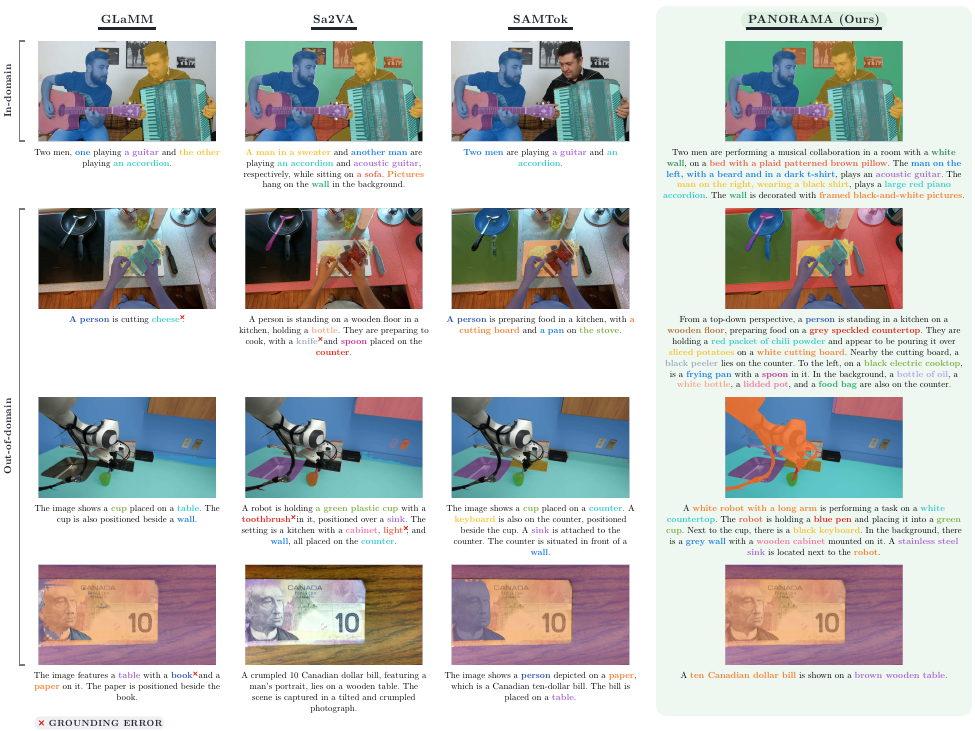}
    \caption{\textbf{Comparison of current models on panoptic grounded captioning.}
    We compare \ourmodel{} with prior grounding models~\citep{rasheed2024glamm, yuan2025sa2va, zhou2026samtok} on an in-domain scene and challenging images from domains not represented in our proposed \ourdataset{}. Prior methods identify relatively few entities and may miss key elements of the scene (e.g., the robot and pen in row 3), segment regions coarsely (e.g., the objects in row 2), and fall back on wrong or generic grounded phrases (e.g., ``cheese'' instead of ``chili powder'' in row 2, ``toothbrush'' for the ``blue pen'' in row 3, or ``paper'' for ``banknote'' in row 4). \ourmodel{} produces denser and more complete descriptions, grounding each mentioned entity with an instance-level mask. Grounded phrases are highlighted in the color of the corresponding predicted mask, and incorrect groundings are marked with a red cross. 
    }
    \label{fig:teaser}
\end{figure}

%% file: sec/02_related_work.tex
\section{Related Work}
\label{sec:related}

\subsection{Spatial Grounding Methods}

\label{sec:related_work_method}

Methods for grounding language outputs in images differ primarily in how they represent spatial information and decode spatial outputs.
One family links generated text to regions using explicit coordinates, learned spatial tokens, or hybrid representations~\citep{chen2023shikra,groma,peng2024kosmos2,you2024ferret}. When localization is restricted to bounding boxes, these approaches provide region-level rather than pixel-level grounding.
A second family couples a multimodal language model with a segmentation module through learned, language-conditioned interfaces. A common design uses the hidden state of a special \texttt{[SEG]} token to condition a dedicated mask decoder that produces the final masks~\citep{zhang2023llavagrounding,chen2024sam4mllm,lai2024lisa,rasheed2024glamm,xia2024gsva,zhang2024omgllava,zhang2024psalm,qian2025reasoningseg,wei2025instructseg,yuan2025sa2va,zhou2024MGLMM,zhang2026evf}.
A more recent direction represents segmentation outputs as discrete codes generated autoregressively~\citep{wang2025argenseg,wang2025himtok,zhou2026samtok}. 
These methods still rely on a learned decoder to reconstruct dense masks, often require an additional stage to train the mask tokenizer, and compress each mask into a fixed number of tokens.
\ourmodel{} adopts a different formulation. For each caption phrase, it conditions a promptable segmenter~\citep{carion2025sam} on the corresponding contextualized VLM representation to generate candidate masks, and then selects the candidates corresponding to the phrase. Unlike GROUNDHOG~\citep{Zhang_2024_CVPR}, which selects masks from an image-level, phrase-independent proposal pool, \ourmodel{} uses this contextualized representation to shape the candidate mask pool before selection. We ablate this choice in \Cref{tab:ablation-selection}. Our formulation leverages the mask quality of a segmentation model trained at scale, requires neither a mask tokenizer nor a separately trained task-specific pixel decoder, and is well suited to phrases that refer to multiple regions.

\subsection{Grounded Captioning Datasets}

\input{figures/dataset_fig}

Prior datasets differ along two dimensions: whether they target comprehension or generation, and whether they provide box- or mask-based grounding. In comprehension tasks, the text is provided, and the goal is to associate textual expressions with corresponding image regions. Flickr30k Entities~\citep{plummer2015flickr30k} links noun phrases to bounding boxes, whereas Panoptic Narrative Grounding~\citep{gonzalez2021panoptic} grounds human narratives~\citep{pont2020localized} in panoptic segments by automatically mapping mouse traces to segmentation regions. These alignments are inferred rather than directly annotated, leaving many phrases without explicit grounding.
By contrast, box-grounded generation datasets require the model to produce a caption while localizing its phrases with bounding boxes~\citep{peng2024kosmos2,lin2025pancap,oliveira2026groundcap}. Such annotations provide only coarse spatial support and cannot precisely represent object boundaries or irregularly shaped non-object regions.
Among mask-grounded generation datasets, GranD-f~\citep{rasheed2024glamm} provides large-scale supervision, but its descriptions are largely repurposed from existing annotations and can be as short as a single phrase, such as \emph{``Snow covered park benches''}. This sparse grounding leaves substantial portions of the scene undescribed, as illustrated on the \emph{left} of \Cref{fig:dataset}.
COCONut-PanCap~\citep{deng2025coconutpancap} pairs dense panoptic masks with grounded captions, but its training images are restricted to COCO, and limited human verification of its automatically generated annotations can leave factual inconsistencies, unreliable phrase-mask alignments, and noisy masks. The \emph{middle} of \Cref{fig:dataset} provides an example of these limitations.
Related resources~\citep{urbanek2024dci, lu2025comprecap} pair masks with human text but target different tasks and do not provide phrase-level alignment.
In contrast, \ourdataset{} provides dense mask annotations paired with human-written captions across diverse images. As illustrated on the \emph{right} of \Cref{fig:dataset}, each referring phrase is explicitly linked to its corresponding image region, enabling detailed supervision and reliable evaluation for panoptic grounded captioning.

%% file: figures/dataset_fig.tex
\begin{figure}[t]
    \centering
    \includegraphics[width=\linewidth]{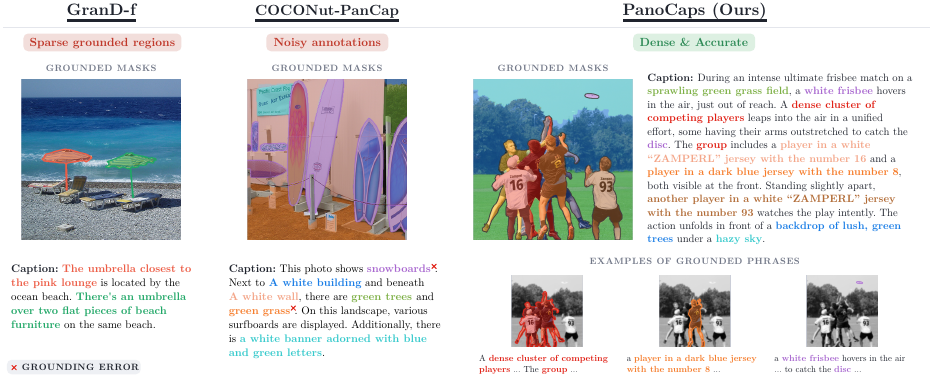}
    \caption{\textbf{\ourdataset{} compared to existing grounded-captioning sources.}
    GranD-f~\citep{rasheed2024glamm} is \emph{sparse}: its captions describe only a few objects, grounding the two umbrellas while leaving the sea, beach, and sunbeds ungrounded.
    COCONut-PanCap~\citep{deng2025coconutpancap} is denser but \emph{noisy}: the description consists largely of generic statements and refers to regions incorrectly (red crosses), mentioning snowboards and describing the dry brown ground as green grass.
    In contrast, \ourdataset{} provides both dense and precise grounding. Its \emph{human-written} captions cover the full scene, including foreground and background elements, with each phrase linked to accurate pixel-level masks.
    The grounded-phrase examples on the right further illustrate \emph{multi-referencing}, \emph{multi-region} grounding, and \emph{free-form} descriptions. 
    }
    \label{fig:dataset}
\end{figure}

%% file: sec/05_dataset.tex
\section{The \ourdataset{} Benchmark}

\input{tables/panocaps_stats}

\label{sec:dataset}

Evaluating and improving models on dense, pixel-level scene understanding requires resources that faithfully measure their ability to describe the full scene and ground each referring phrase to its pixels. Existing grounded captioning datasets fall short of this goal, with incomplete region coverage, limited descriptions and image diversity, and captions and alignments that are often noisy or unreliable due to largely automatic annotation pipelines. We therefore introduce \ourdataset{}, a benchmark for panoptic grounded captioning, constructed through a multi-stage annotation protocol that produces human-written, free-form captions with near-complete region coverage and verified phrase-mask alignments.

\subsection{Annotation Pipeline}

\noindent\textbf{Image Selection.} \ourdataset{} is built from established panoptic segmentation corpora, including COCONut~\citep{deng2024coconut}, ADE20K~\citep{Zhou_2017_CVPR}, and VIPSeg~\citep{Miao_2022_CVPR}. These sources provide complementary image distributions and increase diversity in objects, environments, and visual contexts. We manually curate the image set, retaining only samples with valid panoptic masks and high region coverage. Images with incomplete or incorrect masks are discarded. To avoid split leakage and ensure fair evaluation, training images are drawn only from the source training splits, and validation and test images only from the source evaluation splits. 

\noindent\textbf{Grounded Caption Annotation.}
Each image is annotated with a human-written caption that undergoes independent verification. The text describes the full scene, with every groundable phrase linked to one or more segmentation masks. Annotators write \emph{free-form} natural-language descriptions covering the visible entities and their relationships, and mark the span of each groundable phrase together with the mask or set of masks it denotes. 

\noindent\textbf{Quality Control.}
We adopt a multi-stage annotation and verification process (median $\sim$15~min labeling and $\sim$12~min review per image, $\sim$1{,}530 human hours in total). We first conduct a pilot phase on a subset of images to refine the annotation guidelines and clarify task requirements. Annotators then complete the annotations following the finalized protocol. Finally, \emph{all} annotations undergo independent verification to confirm linguistic quality, completeness of region coverage, and correctness of phrase-mask alignment. Examples that fail verification are revised and re-verified before inclusion, ensuring consistent and reliable annotations suitable for training and evaluation. The full protocol, including the annotation guidelines and interface, is given in \hyperref[app:ann-protocol]{Appendix~\ref*{app:ann-protocol}}.

\input{sec/10_matching}

\subsection{Benchmark Features}

\noindent \textbf{Characteristics.}
As summarized in Table~\ref{tab:panocaps-stats}, \ourdataset{} comprises 3.5K images, approximately 34K panoptic masks, and 31.3K grounded entities across the training, validation, and test splits. Each image contains roughly nine grounded entities, spanning 17.9K unique noun phrases and yielding substantial lexical diversity. The referenced regions cover $\approx 99\%$ of image pixels, encompassing both foreground objects (\emph{things}) and background regions (\emph{stuff}). Importantly, human annotators provide both the captions and the corresponding phrase-mask alignments.
As illustrated in~\Cref{fig:dataset}, each caption is a free-form scene description in which every grounded phrase is linked to the masks it denotes: $9.8\%$ of phrases refer to several masks (e.g., grouping instances of the same category), and $12.5\%$ of masks are referenced by more than one phrase, reflecting how humans naturally describe a scene. Beyond evaluation, the high-fidelity annotations in \ourdataset{} support training and post-training. To our knowledge, \ourdataset{} is the first benchmark to combine human-written free-form captions with near-complete pixel-level phrase grounding. Corpus composition, lexical distribution, additional annotated examples, and a comparison with COCONut-PanCap are provided in \hyperref[app:dataset]{Appendix~\ref*{app:dataset}}.

\noindent \textbf{Evaluation Metrics.} 
We assess models along three axes: caption quality, segmentation accuracy, and grounding quality. Caption quality is measured with CAPTURE~\citep{dong2024benchmarking}, which correlates well with human judgments of detailed captions by parsing candidate and reference captions into sets of objects, attributes, and relations and measuring their agreement. This suits \ourdataset{}, whose captions describe entities and their attributes, and provides a structured measure that complements the grounding evaluation.
Segmentation quality is measured using mask AP at IoU 0.5 (AP50) and mean intersection over union (mIoU), both computed class-agnostically, as \ourdataset{} phrases are free-form and carry no fixed label vocabulary.
Correct grounding requires each predicted entity to be both correctly described and correctly localized, whereas the first two axes assess these capabilities only in isolation. We therefore introduce a matching procedure tailored to the free-form annotations in \ourdataset{}. 
Because phrases are free-form, textual agreement between predicted and ground-truth phrases is determined through a cascade of exact, lemma-based, and WordNet-synonym matching with a sentence-embedding fallback~\citep{miller1995wordnet,reimers2019sentence}. Because a region may be mentioned multiple times or as part of a group, we resolve ambiguous candidate matches before applying Hungarian assignment. The resulting matching procedure is summarized in Procedure~\ref{alg:matching}.
Using the resulting matches, we report Recall following GLaMM~\citep{rasheed2024glamm}, defined as the fraction of ground-truth masks matched by a prediction.
Recall alone, however, can reward over-prediction: a model may recover most ground-truth regions by producing numerous spurious predictions, resulting in cluttered and low-quality outputs. We therefore additionally report Precision and their harmonic mean, F1. These metrics still treat each match as binary and do not capture the degree of agreement between a prediction and its corresponding ground-truth region. To address this, we further introduce a generalized Panoptic Quality (gPQ), which extends standard PQ~\citep{kirillov2019panoptic} from fixed semantic categories to free-form phrases:
\begin{equation}
\mathrm{gPQ} \;=\;
\frac{\displaystyle \sum_{(i,j)\in \mathcal{M}} \text{Sim}_{\text{mask}}(i,j)\,\text{Sim}_{\text{text}}(i,j)}
{|\mathcal{M}| \;+\; \tfrac{1}{2}\,(\mathrm{FP}+\mathrm{FN})}
\label{eqn:gpq}
\end{equation}
where $\mathcal{M}$ is the set of matched pairs, $\text{Sim}_{\text{mask}}$ is the mask IoU, $\text{Sim}_{\text{text}}$ the phrase similarity described above, both in $[0,1]$, and $\mathrm{FP}$ and $\mathrm{FN}$ denote the numbers of unmatched predictions and ground-truth masks, respectively. When textual labels are fixed semantic classes, gPQ reduces to a class-agnostic PQ, pooled over all segments rather than averaged per class, by replacing textual similarity with a binary indicator of same-class agreement. Like F1, gPQ penalizes both unmatched predictions and unmatched ground-truth regions, while additionally weighting matched pairs by their graded textual and mask agreement. \hyperref[app:panocaps_eval]{Appendix~\ref*{app:panocaps_eval}} provides the prediction format, the full metric definitions, the prompt templates, a comparison with the prior grounded conversation generation (GCG) protocol, and a validation of our phrase-similarity measure against human judgments.

%% file: tables/panocaps_stats.tex
\begin{table}[t]
\centering
\scriptsize
\tabcolsep=0.12cm
\setlength{\aboverulesep}{0pt}\setlength{\belowrulesep}{0pt}
\renewcommand{\arraystretch}{1.25}
\caption{\textbf{Statistics of \ourdataset{}.} Coverage denotes the percentage of image pixels covered by panoptic regions grounded by the caption, UNP the number of unique noun phrases, and Things the proportion of masks that are thing rather than stuff regions. / denotes a per-unit average (e.g., entities per image).}
\label{tab:panocaps-stats}
\begin{NiceTabular}{l ccc w{c}{0em} cc w{c}{0em} cc w{c}{0em} c}[colortbl-like]
\CodeBefore
  \rectanglecolor{groupblue}{2-2}{2-4}
  \rectanglecolor{groupblue}{2-6}{2-7}
  \rectanglecolor{groupblue}{2-9}{2-10}
  \rectanglecolor{groupblue}{2-12}{2-12}
\Body
\toprule
 & \multicolumn{3}{c}{\textbf{Size}} & & \multicolumn{2}{c}{\textbf{Density}} & & \multicolumn{2}{c}{\textbf{Language}} & & \textbf{Composition} \\
Split & Images & Masks & Entities & & Entities / img & Coverage (\%) & & Tokens / cap & UNP & & Things (\%) \\
\midrule
Train & 2070 & 21.8K & 19.8K & & 9.6 & 99.3 & & 95.0 & 12.3K & & 64 \\
Test  & 980  & 8.6K  & 7.9K  & & 8.1 & 98.8 & & 82.7 & 5.4K  & & 62 \\
Val   & 420  & 3.7K  & 3.5K  & & 8.3 & 98.7 & & 84.3 & 2.7K  & & 62 \\
\midrule
\rowcolor{rowgray}
\textbf{All}   & 3470 & 34.1K & 31.3K & & 9.0 & 99.1 & & 90.2 & 17.9K & & 63 \\
\bottomrule
\end{NiceTabular}
\end{table}

%% file: sec/10_matching.tex
\setlength{\interspacetitleruled}{0pt}%
\SetAlCapHSkip{0pt}%
\makeatletter
\newcommand{\algocapstrip}[1]{\colorbox{groupblue}{\parbox{\dimexpr\algocf@ruledwidth-2\fboxsep\relax}{\strut #1}}}
\makeatother
\SetAlgoCaptionLayout{algocapstrip}
\SetAlFnt{\footnotesize}
\SetAlCapFnt{\footnotesize}
\SetAlCapNameFnt{\footnotesize}
\SetAlgoNlRelativeSize{-1}
\setlength{\algomargin}{1em}
\begin{algorithm*}[t]
\KwIn{Predicted grounded pairs $P=\{(p^{\text{mask}}_j, p^{\text{text}}_j)\}_{j=1}^m$,
      Ground-truth pairs $G=\{(g^{\text{mask}}_i, g^{\text{text}}_i)\}_{i=1}^n$}
\KwOut{Matched pairs $(g_i, p_j)$ with textual and mask similarity scores;}
\textbf{Normalization.} Lowercase and remove non-informative content (e.g., whitespace) from $p^{\text{text}}$ and $g^{\text{text}}$\;
\textbf{Deduplication.} For each mask in $P$ and in $G$, collect the set of phrases associated with it, giving $D^{P}(j)$ and $D^{G}(i)$. Predicted masks with $\text{IoU}\ge0.9$ whose phrase sets contain a pair with similarity $\ge0.5$ (using $s(\cdot,\cdot)$ defined below) are merged into a single prediction with the union of their phrases\;
\ForEach{ground-truth mask $i$}{
  \ForEach{predicted mask $j$}{
    \tcp{Mask similarity}
    $\text{Sim}_{\text{mask}}(i,j)\gets \text{IoU}(g^{\text{mask}}_i, p^{\text{mask}}_j)$\;
    \tcp{Phrase similarity}
    $\text{Sim}_{\text{text}}(i,j)\gets 0$\;
    \ForEach{$a \in D^{G}(i)$}{
      \ForEach{$b \in D^{P}(j)$}{
        $s(a,b)=1$ if $a=b$, or if $a,b$ are single tokens sharing a WordNet synset on lemmatized forms;
otherwise $s(a,b)=\text{MPNetBaseSim}(a,b)$\;
        $\text{Sim}_{\text{text}}(i,j)\gets \max\{\text{Sim}_{\text{text}}(i,j),\, s(a,b)\}$\;
      }
    }
  }
}
\textbf{Matching.} Run Hungarian matching between predicted and ground-truth masks using the mean similarity 
$\tfrac{1}{2}\big(\text{Sim}_{\text{text}} + \text{Sim}_{\text{mask}}\big)$ and retain matches with $\text{Sim}_{\text{text}}\ge 0.5$ and $\text{Sim}_{\text{mask}}\ge 0.5$.\;
\caption{\textbf{Open-text grounded phrase-mask matching.} Match predicted and ground-truth grounded pairs using textual and mask similarities.}
\label{alg:matching}
\end{algorithm*}

%% file: sec/03_pre.tex
\section{Methodology}
\label{sec:method}

\subsection{Task Definition}

Given an input image $I \in \mathbb{R}^{H \times W \times 3}$, the panoptic grounded captioning task aims to generate a natural language caption $C$ that comprehensively describes the image, while grounding each mentioned entity with a corresponding \emph{set} of masks.
Assume the caption $C$ contains $K$ phrases referring to foreground objects or background regions.
Each phrase may refer to a single region, multiple regions, or be absent from the image.
The goal is to generate comprehensive captions with fine-grained pixel-level grounding, yielding interpretable image descriptions.

%% file: sec/04_method.tex
\input{figures/method}

\subsection{The Proposed Model: \ourmodel{}}

We formulate panoptic phrase grounding as \textbf{selection from a phrase-conditioned pool of mask proposals}. The VLM emits a special \texttt{[SEG]} token per referring phrase, whose final-layer hidden state is projected into a concept vector. This vector conditions a pretrained proposal model to generate candidate masks, from which a learned match scorer selects those corresponding to the phrase.
This design offers three key advantages. First, it leverages the mask quality of a large-scale pretrained segmenter rather than learning mask decoding from scratch. Second, it separates referent identification from boundary delineation: the concept vector encodes the intended referent, while the proposal model produces candidate masks. Third, its set-valued output supports both singular and plural referents while preserving individual instances.
Unlike approaches that collapse multiple instances into a single binary mask~\citep{lai2024lisa, rasheed2024glamm, yuan2025sa2va, zhou2026samtok} or require a separate segmentation token for each instance~\citep{xia2024gsva}, our formulation retains per-instance masks using a single token per phrase. It therefore directly supports the multi-instance grounding annotations in \ourdataset{} while allowing each entity to be individually localized in space.
\Cref{fig:method} illustrates the overall framework, which consists of a vision-language model, a concept bridge, concept-conditioned proposal generation, and proposal selection. We describe each component below.

\noindent \textbf{Vision-Language Model.}
We build on the Qwen3-VL family of vision-language models~\citep{bai2025qwen3}, evaluated at two scales (\Cref{sec:experiments}). The input image $I$ is encoded by the model's native visual encoder and processed together with the user prompt. The VLM autoregressively generates the response, enclosing every grounded phrase within explicit delimiters and emitting one \texttt{[SEG]} token for each grounded phrase: \textit{``\ldots a \texttt{<p>} brown dog \texttt{</p>} \texttt{[SEG]} chasing \texttt{<p>} two frisbees \texttt{</p>} \texttt{[SEG]} \ldots''}.

\noindent \textbf{Concept Bridge.}
Let $\mathbf{h}_k$ denote the final-layer hidden state at the $k$-th \texttt{[SEG]} position, where $k=1,\dots,K$ indexes the grounded phrases. A learned projection $\phi$ maps this state to the prompt space of the proposal model:
\begin{equation}
\mathbf{c}_k = \phi(\mathbf{h}_k) \in \mathbb{R}^{d}, \qquad d = 256 .
\end{equation}
This mapping produces one \textbf{concept vector} per phrase. The concept vector serves as the interface between the VLM and the segmentation branch, conditioning both proposal generation and proposal scoring.

\noindent \textbf{Concept-Conditioned Proposal Generation.}
Our proposal model is built on SAM 3~\citep{carion2025sam}, a pretrained promptable segmenter whose components include a fusion encoder and a DETR-style decoder~\citep{carion2020end} with a segmentation head. In its native form, SAM 3 segments all instances of a short noun phrase encoded by a CLIP-style text encoder~\citep{radford2021clip}. We omit this encoder and instead provide $\mathbf{c}_k$ directly as a single-token prompt. Using the CLIP encoder would condition the segmenter on the phrase text alone, rather than on the VLM's contextualized representation, while also requiring each phrase to be separately encoded by a 354M-parameter text tower. In contrast, the concept token allows the proposal model to be conditioned directly on the VLM representation of the referent, which incorporates both the visual input and preceding linguistic context, while allowing grounding losses to train this representation end to end. We ablate this concept-token design in \Cref{tab:ablation-concept}.
The fusion encoder integrates the concept vector $\mathbf{c}_k$ with the proposal model's image features through cross-attention, producing a concept-conditioned image representation.
The DETR-style decoder operates on this representation and returns a pool of queries for each phrase:
\begin{equation}
\mathcal{Q}_k = \{ \mathbf{q}_{k,1}, \dots, \mathbf{q}_{k,Q} \} .
\end{equation}
Each query $\mathbf{q}_{k,j}$ is a latent embedding from which the segmentation head produces a candidate mask $m_{k,j} \in [0,1]^{H \times W}$. The proposal pool is thus conditioned on the intended referent through $\mathbf{c}_k$, rather than forming a fixed, class-agnostic set.

\noindent \textbf{Proposal Selection.}
A lightweight match scorer compares the concept vector $\mathbf{c}_k$ with each proposal query to estimate how well its associated candidate mask matches the intended referent. Specifically, each query embedding $\mathbf{q}_{k,j}$ is compared with the concept vector through a dot product:

\begin{equation}
s_{k,j} = \sigma\!\left( \big\langle \psi_q(\mathbf{q}_{k,j}),\; \psi_c(\mathbf{c}_k) \big\rangle \right),
\qquad j = 1,\dots,Q ,
\label{eqn:score}
\end{equation}

where $\psi_q$ and $\psi_c$ are learned projections, and $\sigma$ denotes the logistic function. Two properties distinguish our selection mechanism. First, selection is conditioned on the contextual representation of the generated phrase. Second, the match scorer is trained jointly with caption generation using an explicit selection objective. Together with the bridge $\phi$, this objective encourages a representation space in which proposals corresponding to the intended referent receive higher similarity scores than the remaining proposals.
The selected masks for phrase $k$ are those whose scores exceed a fixed threshold $\theta$, applied only at inference:

\begin{equation}
\mathcal{S}_k = \{\, m_{k,j} : s_{k,j} > \theta \,\}.
\label{eqn:select}
\end{equation}
Each proposal is thresholded independently, so $|\mathcal{S}_k|$ may be zero, one, or several, matching the variable cardinality of the target set. The pool itself is phrase-conditioned, and therefore thresholding selects among candidates already specific to the phrase.
We retain the selected masks as separate instance predictions. When a single entity-level binary mask is required by an evaluation protocol, we use their union. Expressions with no target region are handled by the VLM, which emits no \texttt{[SEG]} token.

\subsection{Training and Inference}

\label{sec:method-train}

\noindent \textbf{Proposal Supervision.}
Each grounded phrase $k$ is associated with a ground-truth set of target regions, $\mathcal{G}_k = \{g_{k,1},\dots,g_{k,n_k}\}$. Here, $n_k = 1$ for a stuff region or a singular referent, while $n_k > 1$ for a plural referent.
During training, following the set-prediction formulation of DETR, we match each phrase's target set to its proposals using Hungarian assignment~\citep{kuhn1955hungarian}. The matching cost combines a Dice term, computed using the unthresholded proposal masks, with the match score, weighted by a coefficient $\lambda$:

\begin{equation}
\mathcal{C}(i,j) = 1 - \mathrm{Dice}\!\left(m_{k,j},\, g_{k,i}\right) - \lambda \, s_{k,j}.
\label{eqn:cost}
\end{equation}

Mask overlap provides the primary matching signal, while the score $s_{k,j}$ favors proposals that the match scorer already associates with the intended referent, particularly when several proposals have similar overlap. The resulting assignment provides the positive and negative query labels used by the selection objective. We ablate this score term in \Cref{sec:ablations}.

\noindent \textbf{Objectives.}
We optimize four loss terms:

\begin{equation}
\mathcal{L} = \mathcal{L}_{\text{text}}
            + \mathcal{L}_{\text{sel}}
            + \mathcal{L}_{\text{mask}}
            + \mathcal{L}_{\text{sem}} .
\label{eqn:objective}
\end{equation}
$\mathcal{L}_{\text{text}}$ is the standard next-token cross-entropy loss over the generated response, supervising both the caption text and the special grounding delimiters and \texttt{[SEG]} markers. 
$\mathcal{L}_{\text{sel}}$ is a per-query sigmoid focal loss~\citep{lin2017focal} on the match scores $s_{k,j}$, with binary targets given by the assignment.
$\mathcal{L}_{\text{mask}}$ combines binary cross-entropy and Dice loss between each matched proposal mask $m_{k,j}$ and its assigned target region $g_{k,i}$, providing gradients to the fusion encoder and concept bridge.
Finally, $\mathcal{L}_{\text{sem}}$ applies the same mask losses to the single-channel semantic mask predicted by the proposal model's pretrained semantic head for each concept. The mask is supervised using the union of the target regions, providing assignment-independent supervision to the fusion encoder and concept bridge. The head is discarded at inference, and we ablate this term in \Cref{sec:ablations}.
Thus, the fusion encoder and concept bridge receive gradients from all three grounding losses, while the match scorer is trained only by $\mathcal{L}_{\text{sel}}$. Details of the trainable and frozen components, loss weighting, and optimization hyperparameters are provided in~\Cref{sec:exp_details}.

\noindent \textbf{Inference.}
At inference time, whenever the model emits a \texttt{[SEG]} token, its hidden state is projected to obtain $\mathbf{c}_k$. The proposal model then generates candidate masks, and the match scorer selects those whose scores exceed $\theta = 0.5$, forming $\mathcal{S}_k$ according to \Cref{eqn:select}. The selected masks are upsampled to the input resolution and binarized. Image-backbone features are computed once and shared across all phrases.

\noindent \textbf{A Unified Formulation.}
Our approach unifies existing grounding settings along two independent axes: whether the referring expressions are provided externally or generated by the model, and whether each expression corresponds to a single region or a set of regions. Existing tasks occupy individual points in this space: referring segmentation~\citep{yu2016modeling} assumes a provided expression with a single target instance; generalized referring segmentation~\citep{liu2023gres} extends this to arbitrary target cardinality, including empty sets; and category-level grounding~\citep{hu2025groundingsuite} maps expressions to all matching semantic regions. Grounded captioning introduces the additional challenge of generating the expressions themselves. \ourdataset{} combines both axes: it requires exhaustive scene description while grounding each generated referring phrase to its corresponding panoptic regions.

%% file: figures/method.tex
\begin{figure*}[t]  
    \centering
    \includegraphics[width=\linewidth]{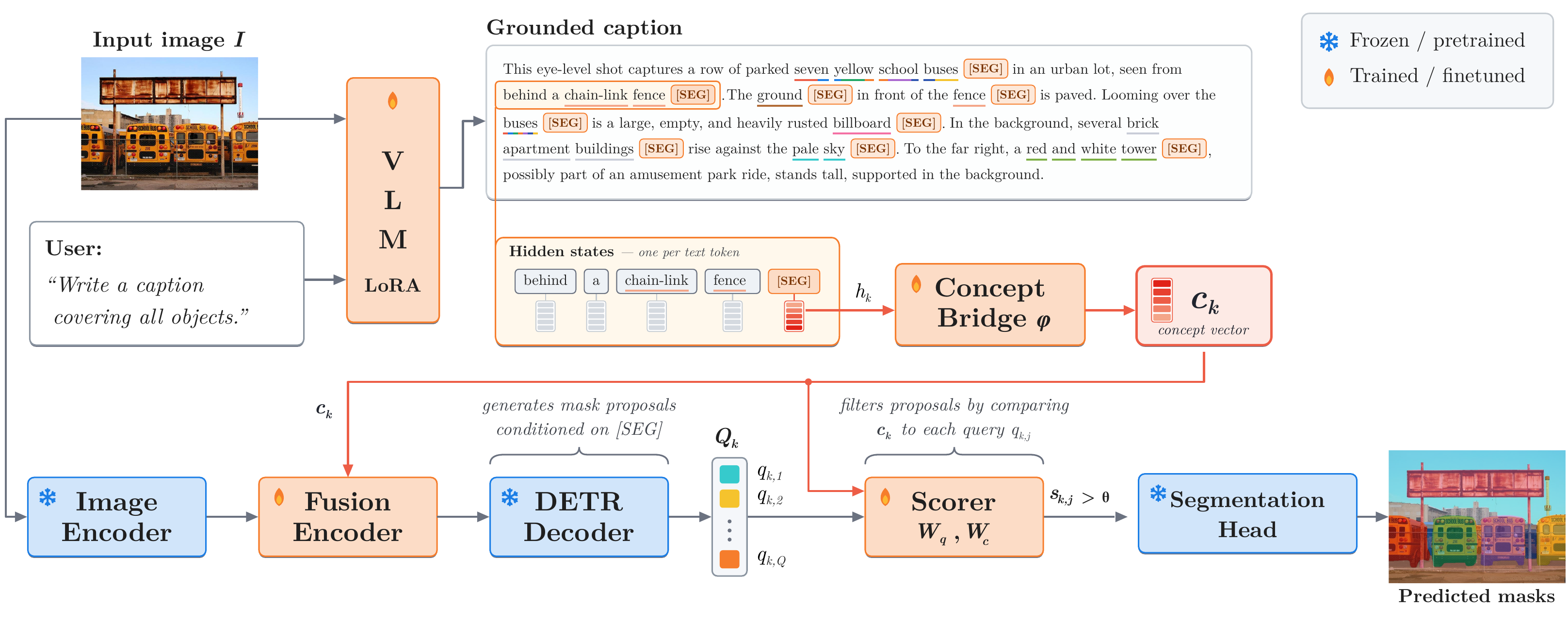} 
    \caption{
        \textbf{Overview of the \ourmodel{} architecture.}
        Given an input image $I$, the \emph{vision-language model} generates a caption containing one \texttt{[SEG]} token for each referring phrase. The final-layer hidden state of each token is projected by the \emph{concept bridge} ($\phi$) into a concept vector ($\mathbf{c}_k$). Conditioned on this vector and the image features, the \emph{proposal model} produces a phrase-specific query pool ($\mathcal{Q}_k$), with each query parameterizing a candidate mask. The \emph{match scorer} compares each query with the concept vector and selects the subset of candidate masks ($\mathcal{S}_k$) associated with the phrase. The selected masks are retained individually, and their union forms the final entity-level prediction. The pipeline is illustrated for a single phrase; the final panel overlays the predicted masks for every phrase in the caption.
    }
    \label{fig:method}
\end{figure*}

%% file: sec/06_abla.tex
\section{Experiments}

\label{sec:experiments}

\input{tables/mixture}

\subsection{Experimental Setting}
\label{sec:exp_details}

\noindent \textbf{Training Data.}
To develop strong scene-understanding and grounded-generation capabilities, we construct a 957K-sample training mixture spanning the tasks described above. Dense grounded captioning forms the largest component, combining the scale of COCONut-PanCap~\citep{deng2025coconutpancap} with the high-quality, human-verified supervision of \ourdataset{}.
COCONut-PanCap contains noisy captions and missing phrase-mask correspondences. We therefore preprocess its samples before incorporating them into the training mixture. Specifically, we use Qwen3-VL-235B~\citep{bai2025qwen3} to regenerate grounded captions, prompting the model to describe each image while associating each mentioned object with its corresponding panoptic mask and object ID. We then apply a second verification stage using Qwen3-VL and automated parsing tools to identify and correct issues, including incorrect object assignments, redundant caption phrases or object tags, and malformed or invalid object references.
The remaining groups differ in whether the grounded phrases are generated by the model or provided as input, and whether each phrase refers to a single region or multiple regions. Table~\ref{tab:mixture} summarizes the composition and sampling proportions of the resulting mixture.
The multi-granularity group is adapted from SegLLM~\citep{wang2025segllm}. Because its original samples provide target regions as input, they are incompatible with our mask-free input setting. We therefore use Qwen3-VL to retain samples whose targets can be identified from the image and text alone. This group adds parts, stuff regions, attributes, and relationally specified regions, broadening the diversity of targets encountered during training.

\noindent \textbf{Implementation Details.}
We implement \ourmodel{} in XTuner~\citep{2023xtuner} based on the Qwen3-VL-Instruct family~\citep{bai2025qwen3}. We evaluate the 4B variant and use the 2B model for ablations. 
The proposal model receives images resized to a longest side of 1008 pixels, while the VLM operates at its native dynamic resolution. We keep the segmenter's native proposal pool of $Q=200$ queries.
We adapt the language model with LoRA~\citep{hu2022lora} on all attention and MLP projections, using dropout 0.05, rank 256 and $\alpha{=}512$ for the 4B model, and rank 128 and $\alpha{=}256$ for the 2B model. We additionally train the token embeddings and output head, the concept bridge $\phi$, and the proposal model's fusion encoder and match scorer. All other parameters, including both vision encoders, remain frozen. This choice is ablated in \Cref{sec:ablations}.

In \Cref{eqn:objective}, the mask, semantic, and language losses contribute with unit weight, and the selection loss with weight $0.5$; within the mask losses, the BCE and Dice terms are weighted by $2.0$ and $0.5$, following~\citet{yuan2025sa2va}. Sweeping these weights within the ranges we explored did not strongly affect the results. We use $\lambda{=}0.25$ in \Cref{eqn:cost} and select proposals with $s_{k,j}>\theta{=}0.5$ at inference. We ablate this threshold and compare it with selecting a fixed number of proposals per phrase in \hyperref[app:selection-rule]{Appendix~\ref*{app:selection-rule}}. At inference, captions are generated greedily. During training, we supervise at most 10 grounded phrases per image, uniformly sampling them when an image contains more. The cap bounds peak memory and keeps the number of mask targets per step uniform across images, and lies above the average of 9.6 grounded entities per training image reported in \Cref{tab:panocaps-stats}. Unless otherwise stated, results are from a single training run. To quantify run-to-run variability, we train the 2B model with three seeds, the scale at which repeated training is practical, obtaining standard deviations of 0.6 gPQ, 1.0 AP50, and 0.1 mean RefCOCO cIoU; differences of this magnitude should be read as comparable performance.

We train for one epoch on the full mixture with AdamW using a learning rate of $4\times10^{-5}$, weight decay $0.05$, and gradient clipping at $1.0$. We apply linear warmup over the first $5\%$ of steps, followed by cosine decay to zero, and use BF16 precision with a maximum sequence length of 8,192 tokens. We use an effective batch size of 128. All models are trained on 16 NVIDIA H100 GPUs. Training on the full mixture takes approximately 5 and 7 wall-clock hours for the 2B and 4B models, respectively. Per-dataset finetuning starts from the mixture-trained checkpoint and uses the same configuration on each benchmark's training split. Baseline methods are finetuned on \ourdataset{} for 10 epochs using the released code and hyperparameters and the \ourdataset{} prompt. For baselines that predict a single mask per phrase, phrases grounded to multiple regions are supervised with the union of their ground-truth masks. Zero-shot evaluations use each model's GCG prompt, the closest task to panoptic grounded captioning on which these models were trained. Trainable and total parameter counts, inference latency, and peak memory for \ourmodel{} and all finetuned baselines are reported in \hyperref[app:efficiency]{Appendix~\ref*{app:efficiency}}.

\subsection{Ablation Studies}
\label{sec:ablations}

We conduct extensive ablations to isolate the contributions of the proposed architecture and \ourdataset{}. Unless otherwise stated, all architectural ablations use the 2B model and keep the training data, optimization schedule, and effective batch size fixed.

\input{tables/ablation_selection}
\input{tables/ablation_concept}

\noindent \textbf{Direct Mask Generation.}
We first examine the central design choice of \ourmodel{}: using the \texttt{[SEG]} token to select from a pool of mask proposals rather than directly decoding a mask. Prior grounded VLMs, including GLaMM~\citep{rasheed2024glamm} and Sa2VA~\citep{yuan2025sa2va}, typically use the \texttt{[SEG]} representation to directly decode the target mask. 
We therefore construct a matched direct-generation variant, reported as \emph{w/o selection} in \Cref{tab:ablation-selection}, that uses the same VLM, training data, and optimization schedule, but projects the \texttt{[SEG]} hidden state into SAM 3's promptable mask decoder. The decoder is trained to adapt to the projected representation, so that the two models differ mainly in how masks are decoded.
Direct mask generation leaves captioning unchanged and referring segmentation essentially intact, yet costs 7.5 gPQ and 20.0 AP50 on \ourdataset{}, so the deficit is concentrated in dense grounding. Replacing this decoder with that of SAM 2, used by prior work, changes gPQ by only 0.4, within seed variation, so the gap is not specific to the choice of decoder.
At a matched training budget, the result suggests that dense grounding benefits from separating \emph{referent identification} from \emph{mask generation}. In direct generation, the \texttt{[SEG]} representation must simultaneously encode the linguistic identity of the referent and provide the information needed for pixel-level mask prediction. Our model instead uses the concept representation to score and select candidate masks, allowing the segmentation module to handle the low-level spatial details of mask construction. 

\noindent \textbf{Phrase-Conditioned Proposals.}
A further difference from prior selection-based methods is that the proposal pool is conditioned on the phrase, whereas methods such as GROUNDHOG~\citep{Zhang_2024_CVPR} select from a single pool generated once per image. We test this with the \emph{w/o proposal conditioning} variant in \Cref{tab:ablation-selection}, which generates one image-level pool and lets every phrase select from it. Removing the conditioning costs 1.8 gPQ and 2.6 Recall on \ourdataset{}, and 2.9 RefCOCO cIoU. Conditioning steers proposal generation toward the entity the phrase names, and the drop is accordingly largest on referring segmentation, where that entity must be singled out among several same-category instances.

\noindent \textbf{Score-Aware Proposal Matching.}
We further ablate the score-aware proposal matching by removing the score term from the Hungarian assignment cost, setting $\lambda=0$ in \Cref{eqn:cost}. 
During training, Hungarian assignment determines which proposal serves as the positive target for each ground-truth region, and the score term favors, among proposals with similar mask overlap, the one the match scorer associates with the referent.
In the \emph{w/o score-aware matching} row of \Cref{tab:ablation-selection}, relying only on mask overlap reduces gPQ by 2.4 points and AP50 by 7.5 points on \ourdataset{}. This indicates that when several proposals overlap a target similarly, the match score provides a useful additional signal for choosing which one to supervise.

\noindent \textbf{Dense Semantic Supervision.}
We ablate the dense semantic supervision $\mathcal{L}_{\text{sem}}$ in \Cref{tab:ablation-selection}. This loss supervises the segmenter's single-channel semantic mask, which is used only during training. Removing it reduces average RefCOCO cIoU by 2.5 points, while \ourdataset{} remains within seed variation. Because it supervises a single mask covering the whole referent, independently of proposal assignment, this signal benefits single-entity grounding, where the referent must be segmented as one region. We therefore retain it in the default model.

\noindent \textbf{Concept Source.}
We study the source of the concept representation in \Cref{tab:ablation-concept}.
Conditioning proposal generation on a phrase embedding from a CLIP-style text encoder, as in the \emph{CLIP text prompt} variant, reduces RefCOCO cIoU by 4.1 points. Re-introducing the projected \texttt{[SEG]} representation into the CLIP prompt, as in the \emph{CLIP + \texttt{[SEG]} prompt} variant, recovers the RefCOCO gap and matches the default within seed variation on every metric. This underscores the importance of conditioning proposal generation on the contextualized VLM representation, which carries grounding information that the phrase text embedding alone does not.
Using the projected \texttt{[SEG]} representation also allows removing the CLIP-style text encoder entirely: \ourmodel{} uses this representation for both roles of the encoder, proposal conditioning and scoring, which saves approximately 354M parameters and a per-phrase text-encoding pass. 
A natural question is whether the concept vector $\mathbf{c}_k$, a single projection of the \texttt{[SEG]} hidden state, discards textual content that the proposal model could use.
In the \emph{w/ phrase span tokens} variant, we instead condition the proposal model on the projected hidden states of all tokens in the phrase, together with $\mathbf{c}_k$. This leaves every metric within seed variation, indicating that $\mathbf{c}_k$ already summarizes the phrase content.

\noindent \textbf{Trainable Components.}
In the default configuration, only the fusion encoder and match scorer are trained, together with the concept bridge and the VLM via LoRA. We vary this by either freezing the proposal model entirely or additionally unfreezing the DETR decoder and mask heads. Neither modification improves performance: \ourdataset{} metrics remain within seed variation, and both slightly reduce RefCOCO cIoU. This suggests that performance is driven primarily by learning the interface between the VLM and proposal model rather than by adapting the pretrained proposal model itself. \hyperref[app:ablation-trainable]{Appendix~\ref*{app:ablation-trainable}} reports these experiments in full.

\input{tables/ablation_data_gcg}

\noindent \textbf{Dataset Contribution.}
We next examine the contribution of \ourdataset{} to panoptic grounded captioning. \Cref{tab:abl-data-gcg} compares models trained on a single grounded captioning source at a time. Each source is matched to the same training budget using repeat factors, so that differences reflect the training data rather than the number of optimization steps. For COCONut-PanCap, we use the regenerated captions described in \Cref{sec:exp_details} rather than the released annotations, so the comparison is made against a cleaned version of this source. \ourdataset{} contains only 2,070 training images, $57\times$ fewer than COCONut-PanCap~\citep{deng2025coconutpancap} and $95\times$ fewer than the GCG training samples~\citep{rasheed2024glamm}, but provides human-written, fully verified, free-form captions with dense phrase-mask alignments. Despite its smaller scale, \ourdataset{} achieves the strongest overall performance on its own benchmark, improving gPQ by 6.1 and 25.2 points, respectively. The gain over COCONut-PanCap is not a segmentation effect: both models achieve comparable segmentation quality, while Recall increases from 52.0 to 65.5 and CAPTURE from 64.7 to 67.8. The advantage instead appears as broader coverage of scene entities and higher caption quality. On GCG, which is in-domain for the GCG training data but out-of-domain for both dense sources, training on \ourdataset{} transfers better than training on COCONut-PanCap on every metric ($+5.8$ Recall, $+5.6$ mIoU, $+4.0$ METEOR, and $+2.5$ AP50). These results suggest that, for dense grounded captioning, annotation coverage and alignment fidelity can substantially offset dataset scale, with the benefit extending beyond our own benchmark.

%% file: tables/mixture.tex
\begin{table}[t]
\centering
\scriptsize
\tabcolsep=0.12cm
\caption{\textbf{\ourmodel{} training mixture.} Composition by capability group, reported as the number and proportion of training samples. For each group, we indicate whether phrases are generated by the model or provided as input (\emph{Phrases}), the number of regions grounded per phrase (\emph{Regions/phrase}), and the contributing datasets (\emph{Sources}).
}
\label{tab:mixture}
\begin{NiceTabular}{clll>{\raggedright\arraybackslash}m{5cm}rr}
\CodeBefore
  \rectanglecolor{groupblue}{2-1}{3-1}   %
  \rectanglecolor{groupblue}{4-1}{6-1}   %
\Body
\toprule
& Capability & Phrases & Regions/phrase & Sources & Samples & Share \\ \midrule
\multirow{2}{*}{\rotatebox[origin=c]{90}{Caption}}
& \textbf{Dense grounded captioning}          & generated & one or more       & \textbf{\ourdataset{}}, \mbox{COCONut-PanCap~\citep{deng2025coconutpancap}} & \textbf{325{,}062} & \textbf{34.0\%} \\ \addlinespace[4pt]
& Grounded conversation generation   & generated & one or more       & GranD-f, Flickr30k, OpenPSG, RefCOCOg~\citep{rasheed2024glamm} & 171{,}596 & 17.9\% \\ \midrule
\multirow{3}{*}[-12pt]{\rotatebox[origin=c]{90}{Segmentation}}
& Referring segmentation             & given     & exactly one       & RefCOCO/+/g~\citep{yu2016modeling} & 257{,}526 & 26.9\% \\ \addlinespace[4pt]
& Generalized referring segmentation & given     & zero, one or more & gRefCOCO~\citep{liu2023gres}, PhraseCut~\citep{wu2020phrasecut} & 101{,}866 & 10.6\% \\ \addlinespace[4pt]
& Multi-granularity segmentation     & given     & exactly one       & PACO~\citep{ramanathan2023paco}, PASCAL-Part~\citep{mottaghi2014role}, LVIS~\citep{gupta2019lvis}, VG~\citep{krishna2017visual}, COCO-Stuff~\citep{caesar2018coco}, ADE20K~\citep{Zhou_2017_CVPR} & 101{,}349 & 10.6\% \\ \midrule
& Total                              &           &                   & & 957{,}399 & 100\% \\
\bottomrule
\end{NiceTabular}
\end{table}

%% file: tables/ablation_selection.tex
\providecommand{\sd}[1]{{\tiny$\pm$#1}}
\providecommand{\dt}[1]{\,{\color{gray}\scriptsize(#1)}}
\providecommand{\dtb}[1]{\,{\scriptsize\textbf{(#1)}}}
\begin{table}[t]
\centering
\scriptsize
\tabcolsep=0.15cm
\setlength{\aboverulesep}{0pt}\setlength{\belowrulesep}{0pt}
\renewcommand{\arraystretch}{1.25}
\caption{\textbf{Component ablations for \ourmodel{}-2B.} Each variant is evaluated on the \ourdataset{} validation split and modifies a single component of the default model while keeping the training data, schedule, and effective batch size fixed. RefCOCO reports mean cIoU over the three validation splits. Every configuration is reported as the mean of three training seeds and $\pm$ is the standard deviation across them. Gray values denote changes relative to the default; \textbf{bold} marks statistically significant differences from the default (two-sample $t$-test, $p<0.05$).
}
\label{tab:ablation-selection}
\resizebox{\linewidth}{!}{%
\begin{tabular}{lcccccc}
\toprule
\textbf{Configuration} & gPQ & Recall & Precision & AP50 & CAPTURE & RefCOCO \\
\midrule
\rowcolor{groupblue}
\ourmodel{}-2B (default) & 42.1\sd{0.6} & 60.3\sd{0.5} & 63.9\sd{1.2} & 60.8\sd{1.0} & 66.2\sd{0.2} & 80.4\sd{0.1} \\
\midrule
\quad w/o selection (SAM 3 decoder) & 34.6\sd{0.8}\dtb{-7.5} & 50.3\sd{0.5}\dtb{-10.0} & 54.3\sd{1.7}\dtb{-9.6} & 40.8\sd{1.3}\dtb{-20.0} & 66.3\sd{0.1}\dt{+0.1} & 80.1\sd{0.0}\dtb{-0.3} \\
\quad w/o selection (SAM 2 decoder) & 35.0\sd{0.5}\dtb{-7.1} & 50.6\sd{0.4}\dtb{-9.7} & 55.9\sd{0.9}\dtb{-8.0} & 42.5\sd{0.6}\dtb{-18.3} & 66.4\sd{0.1}\dt{+0.2} & 79.9\sd{0.1}\dtb{-0.5} \\
\quad w/o proposal conditioning & 40.3\sd{0.5}\dtb{-1.8} & 57.7\sd{0.5}\dtb{-2.6} & 63.5\sd{0.9}\dt{-0.4} & 59.2\sd{1.6}\dt{-1.6} & 65.8\sd{0.2}\dt{-0.4} & 77.5\sd{0.2}\dtb{-2.9} \\
\quad w/o score-aware matching & 39.7\sd{0.5}\dtb{-2.4} & 53.5\sd{0.7}\dtb{-6.8} & 64.5\sd{0.8}\dt{+0.6} & 53.3\sd{0.5}\dtb{-7.5} & 66.0\sd{0.2}\dt{-0.2} & 80.1\sd{0.0}\dtb{-0.3} \\
\quad w/o $\mathcal{L}_{\text{sem}}$ & 42.6\sd{0.3}\dt{+0.5} & 61.1\sd{0.6}\dt{+0.8} & 65.3\sd{0.4}\dt{+1.4} & 61.1\sd{0.7}\dt{+0.3} & 66.5\sd{0.3}\dt{+0.3} & 77.9\sd{0.1}\dtb{-2.5} \\
\bottomrule
\end{tabular}
}
\end{table}

%% file: tables/ablation_concept.tex
\providecommand{\sd}[1]{{\tiny$\pm$#1}}
\providecommand{\dt}[1]{\,{\color{gray}\scriptsize(#1)}}
\providecommand{\dtb}[1]{\,{\scriptsize\textbf{(#1)}}}
\begin{table}[t]
\centering
\scriptsize
\tabcolsep=0.15cm
\setlength{\aboverulesep}{0pt}\setlength{\belowrulesep}{0pt}
\renewcommand{\arraystretch}{1.25}
\caption{\textbf{Concept representation ablations for \ourmodel{}-2B.}
Variants differ only in how concept representations are provided to the model. Results are averaged over three training seeds, with $\pm$ the standard deviation across them; gray values show changes relative to the default, with \textbf{bold} marking statistically significant differences from the default ($p<0.05$, three seeds).
}
\label{tab:ablation-concept}
\resizebox{\linewidth}{!}{%
\begin{tabular}{lcccccc}
\toprule
\textbf{Configuration} & gPQ & Recall & Precision & AP50 & CAPTURE & RefCOCO \\
\midrule
\rowcolor{groupblue}
\ourmodel{}-2B (default) & 42.1\sd{0.6} & 60.3\sd{0.5} & 63.9\sd{1.2} & 60.8\sd{1.0} & 66.2\sd{0.2} & 80.4\sd{0.1} \\
\midrule
\quad CLIP text prompt & 41.6\sd{0.8}\dt{-0.5} & 59.4\sd{1.0}\dt{-0.9} & 64.3\sd{1.3}\dt{+0.4} & 61.3\sd{0.7}\dt{+0.5} & 66.0\sd{0.1}\dt{-0.2} & 76.3\sd{0.2}\dtb{-4.1} \\
\quad CLIP + \texttt{[SEG]} prompt & 42.2\sd{0.4}\dt{+0.1} & 60.3\sd{0.8}\dt{0.0} & 64.3\sd{0.3}\dt{+0.4} & 60.9\sd{1.3}\dt{+0.1} & 66.1\sd{0.2}\dt{-0.1} & 80.3\sd{0.1}\dt{-0.1} \\
\quad w/ phrase span tokens & 42.6\sd{0.5}\dt{+0.5} & 61.1\sd{0.4}\dt{+0.8} & 64.7\sd{0.9}\dt{+0.8} & 61.0\sd{0.8}\dt{+0.2} & 66.4\sd{0.2}\dt{+0.2} & 80.4\sd{0.2}\dt{0.0} \\
\bottomrule
\end{tabular}
}
\end{table}

%% file: tables/ablation_data_gcg.tex
\begin{table}[t]
\centering
\scriptsize
\tabcolsep=0.08cm
\setlength{\aboverulesep}{0pt}\setlength{\belowrulesep}{0pt}
\renewcommand{\arraystretch}{1.25}
\caption{\textbf{Training-data ablation for \ourmodel{}-2B.} Each model is trained on a single grounded captioning source under the same training budget, using repeat factors to match the number of training instances, and is evaluated on the \ourdataset{} and GCG~\citep{rasheed2024glamm} validation splits. Scale denotes the source's number of unique training samples relative to \ourdataset{}. $^{*}$ marks our regenerated annotations. The GCG row combines GLaMM's four GCG training sources (GranD-f, Flickr30k, OpenPSG, and RefCOCOg), so GCG evaluation is in-domain for it but out-of-domain for the two dense sources. Despite using 57$\times$ fewer samples, \ourdataset{} achieves the strongest overall performance on PanoCaps and transfers better than COCONut-PanCap on zero-shot GCG.
}

\label{tab:abl-data-gcg}
\begin{NiceTabular}{lccccccw{c}{0em}cccc}[colortbl-like]
\CodeBefore
  \rectanglecolor{groupblue}{2-3}{2-7}   %
  \rectanglecolor{groupblue}{2-9}{2-12}  %
\Body
\toprule
 & & \multicolumn{5}{c}{\ourdataset{} (val)} & & \multicolumn{4}{c}{GCG (val)} \\
\cmidrule(lr){3-7}\cmidrule(lr){9-12}
\textbf{Train Data} & Scale & gPQ & Recall & AP50 & mIoU & CAPTURE & & Recall & AP50 & mIoU & METEOR \\
\midrule
 GCG~\citep{rasheed2024glamm} & $\times$95 & 18.8 & 20.5 & 29.6 & 39.8 & 42.7 & & 46.7 & \textbf{36.9} & \textbf{71.5} & 16.1 \\
 COCONut-PanCap$^{*}$~\citep{deng2025coconutpancap} & $\times$57 & 37.9 & 52.0 & \textbf{57.0} & 64.6 & 64.7 & & 42.3 & 26.9 & 64.9 & 13.5 \\
\rowcolor{rowgray}
 \ourdataset{} (Ours) & $\times$1 & \textbf{44.0} & \textbf{65.5} & \textbf{57.0} & \textbf{64.7} & \textbf{67.8} & & \textbf{48.1} & 29.4 & 70.5 & \textbf{17.5} \\
\bottomrule
\end{NiceTabular}
\end{table}

%% file: sec/07_expr.tex
\subsection{Comparison with the State of the Art}

We compare \ourmodel{} with specialized grounding models and generalist VLMs on \ourdataset{} and established grounding benchmarks. We first analyze panoptic grounded captioning, the task targeted by our benchmark, and then assess how the resulting capabilities transfer to grounded conversation generation, referring expression segmentation, and generalized grounding.

\subsubsection{Panoptic Grounded Captioning}
We evaluate panoptic grounded captioning on \ourdataset{}, with results reported in Table~\ref{tab:panocaps}. We consider two model families: specialized grounding models and generalist vision-language models. For specialized models, we report both zero-shot performance using publicly available checkpoints and results after finetuning on \ourdataset{}. As generalist baselines, we evaluate the closed-source Gemini~2.5~Pro API~\citep{comanici2025gemini} and the open-source Qwen3-VL-235B-A22B model~\citep{bai2025qwen3}.

Because generalist VLMs cannot directly produce densely grounded captions, we adapt their inference procedures to our setting. For both Gemini~2.5~Pro and Qwen3-VL, we first prompt the model to generate a caption with the entities to be grounded. For Gemini~2.5~Pro, we issue a second prompt to localize the corresponding instances. Since Qwen3-VL produces bounding-box grounding outputs, we convert the predicted boxes into masks using SAM. Further implementation details are provided in \hyperref[app:generalist-prompt]{Appendix~\ref*{app:generalist-prompt}}.

\input{tables/sota_panocaps}

Generalist VLMs transfer their strong captioning ability to our benchmark, but their pixel-level localization remains substantially less accurate (\emph{top} section of Table~\ref{tab:panocaps}). Gemini~2.5~Pro and Qwen3-VL achieve high captioning scores and moderate grounding recall, indicating that they identify many relevant entities. However, their segmentation quality remains limited, yielding gPQ scores of only 22.7 and 26.1, respectively. Thus, broad entity coverage alone does not translate into precise panoptic region delineation.

Zero-shot specialized models exhibit the opposite limitation (\emph{middle} section of Table~\ref{tab:panocaps}). Optimized primarily for grounding short referring expressions, they retain reasonable segmentation quality but provide limited scene descriptions. Their low grounding recall indicates that they ground only a small fraction of the annotated entities, as illustrated in \Cref{fig:teaser}. Neither model family satisfies the joint requirement of \ourdataset{}: comprehensive scene description together with precise pixel-level grounding.

Training on \ourdataset{} (\emph{bottom} section of Table~\ref{tab:panocaps}) substantially improves specialized models across captioning, segmentation, and grounding. SAMTok-8B, the strongest baseline, improves from 44.6 to 70.6 CAPTURE, from 23.0 to 67.1 Recall, and from 19.7 to 44.5 gPQ. Across architectures, finetuning improves gPQ by 15--25 points, highlighting the value of \ourdataset{}'s dense, panoptic, human-verified supervision.

\ourmodel{}-4B achieves the best results on five of the seven reported metrics. The finetuned baselines retain an edge on the text-only CAPTURE score (up to 70.6 vs.\ 68.0), and the SAMTok models marginally on grounding recall (up to 67.1 vs.\ 66.4). The advantage of \ourmodel{}-4B comes from more accurate pixel-level grounding of generated descriptions: at comparable grounding recall, it achieves higher precision and mask quality, and the best gPQ (45.6 vs.\ 44.5). Notably, \ourmodel{}-4B also outperforms the generalist VLM baselines across all metrics, despite using substantially fewer parameters.

\subsubsection{Additional Results}

\noindent \textbf{Grounded Conversation Generation.}
We evaluate \ourmodel{} on the GCG benchmark~\citep{rasheed2024glamm}, which requires models to generate an image description interleaved with region masks for the entities mentioned in the text. Compared with \ourdataset{}, GCG contains shorter descriptions that typically cover fewer entities. As shown in Table~\ref{tab:gcg}, \ourmodel{}-4B matches or exceeds the strongest specialized models: after finetuning, it obtains the best result on seven of the ten reported metrics, including every grounding metric on both splits. The remaining gaps are in caption quality, where SAMTok-7B (ft) leads by 0.5 METEOR and 0.6 CIDEr on validation and by 0.2 METEOR on test. The dense grounding capabilities developed through training with \ourdataset{} therefore transfer to the shorter and sparser GCG setting.

\input{tables/sota_GCG}

\noindent \textbf{Referring Expression Segmentation.}
This task requires identifying and segmenting the single region specified by a natural-language expression. We assess \ourmodel{} on RefCOCO, RefCOCO+, and RefCOCOg, with results reported in Table~\ref{tab:refcoco}. Without any benchmark-specific finetuning, \ourmodel{}-4B already matches the strongest specialized model, leading SAMTok-4B on four of the eight splits; finetuning adds at most 1.5 cIoU and gives the best result on seven of the eight, with SAMTok-4B remaining best on RefCOCO+ val (80.2 vs.\ 79.8 cIoU).  We also compare against the full SAM 3 Agent, which pairs SAM 3 with a generalist VLM. The SAM 3 Agent built on Gemini~2.5~Pro~\citep{carion2025sam} trails \ourmodel{}-4B by 7--12 cIoU across splits, underscoring the importance of learned alignment between contextual phrase representations and the segmenter's proposal space. Overall, these results show that \ourmodel{} achieves strong single-region grounding despite being designed for more general, densely grounded outputs.

\input{tables/sota_refcoco}
\input{figures/examples}

\noindent \textbf{Generalized Grounding.}
We further evaluate \ourmodel{} on GRES and GroundingSuite. GRES includes expressions that may refer to one or more objects, as well as expressions with no valid target. As reported in Table~\ref{tab:gres}, \ourmodel{} achieves the highest no-target accuracy (N-acc) and the best region-level segmentation, as measured by gIoU and cIoU, across all splits. We also evaluate on GroundingSuite, which covers a broader range of entity types and referring patterns. No GroundingSuite data is used to train \ourmodel{} or any compared model, so the benchmark directly tests generalization to unseen expressions and entity types. As shown in Table~\ref{tab:grounding-suite}, \ourmodel{} ranks first in every category and improves overall gIoU from 67.8 to 75.5. The largest gains appear in the \emph{Part} and \emph{Multi} categories, where gIoU improves from 40.8 to 53.5 and from 72.1 to 80.0, respectively, indicating that supervision from related grounding tasks transfers to the new benchmark. The \emph{Multi} gain reflects a key strength of our architecture: each phrase can select several proposals, which allows multi-region grounding. Together, these results demonstrate that \ourmodel{} generalizes beyond single-region grounding to expressions with multiple or no valid targets.

\input{tables/sota_GRES}
\input{tables/sota_groundingsuite}

\noindent \textbf{Qualitative Examples.}
\Cref{fig:teaser} illustrates panoptic grounded captioning by comparing \ourmodel{} with prior grounding models on one in-domain scene and three images from domains absent from \ourdataset{}. While previous approaches often omit relevant entities, provide incomplete descriptions, or produce coarse and inaccurate masks, \ourmodel{} generates more complete and coherent captions while precisely grounding each mentioned entity with pixel-level masks. \Cref{fig:examples} illustrates the breadth of \ourmodel{} across its evaluated tasks, including panoptic grounded captioning on \ourdataset{}, grounded conversation generation, and generalized grounding. Together, these examples demonstrate the precision and versatility of \ourmodel{}'s grounding capabilities across tasks and domains.

\noindent \textbf{Limitations and Failure Cases.}
Our work focuses on image-level grounding and does not yet cover related settings such as region description or video grounding. The qualitative examples also highlight opportunities for further improvement: \ourmodel{} may occasionally omit visible regions (e.g., wall outlets in \Cref{fig:teaser}, row 3) and may produce imprecise descriptions (e.g., lidded pot, row 2). A feedback loop in which the VLM refines its predictions based on the segmenter's outputs could improve coverage and reduce such errors. Reinforcement learning could further optimize grounding and description quality. Extending the framework to video and region-level description, while exploring these directions, is a natural avenue for future work.

%% file: tables/sota_panocaps.tex
\begin{table}[t]
\centering
\scriptsize
\tabcolsep=0.08cm
\setlength{\aboverulesep}{0pt}\setlength{\belowrulesep}{0pt}
\renewcommand{\arraystretch}{1.25}
\caption{\textbf{Results on \ourdataset{}, our panoptic grounded captioning benchmark.}
Generalist VLMs are evaluated using a two-stage captioning-grounding protocol, while specialized grounding models are evaluated either zero-shot or after finetuning on \ourdataset{}. All results are on the \ourdataset{} test split.
Finetuning with \ourdataset{} substantially improves grounding quality, enabling state-of-the-art panoptic grounded captioning performance.
Best and second-best results are shown in \textbf{bold} and \underline{underline}, respectively.
}
\label{tab:panocaps}
\begin{tabular}{lccccccc} \toprule
 & \multicolumn{1}{c}{\textbf{Captioning}} & \multicolumn{2}{c}{\textbf{Segmentation}} & \multicolumn{4}{c}{\textbf{Grounding}} \\
\cmidrule(lr){2-2} \cmidrule(lr){3-4} \cmidrule(lr){5-8}
Method & CAPTURE & AP50 & mIoU & Recall & Precision & F1 & gPQ \\
\midrule
\rowcolor{groupblue}
\multicolumn{8}{c}{\emph{Generalist VLMs}} \\ \addlinespace[3pt]
Gemini 2.5 Pro~\citep{comanici2025gemini} & 64.7 & 25.2 & 35.1 & 50.9 & 30.2 & 37.9 & 22.7 \\
Qwen3-VL-235B-A22B~\citep{bai2025qwen3} & 64.0 & 30.7 & 44.5 & 52.7 & 36.4 & 43.1 & 26.1 \\
\midrule
\rowcolor{groupblue}
\multicolumn{8}{c}{\emph{Specialized grounding models without training on \ourdataset{}}} \\ \addlinespace[3pt]
OMG-LLaVA-7B~\citep{zhang2024omgllava} & 39.0 & 21.1 & 33.8 & 15.7 & 45.4 & 23.3 & 14.9 \\
GLaMM-7B~\citep{rasheed2024glamm} & 40.7 & 22.3 & 34.8 & 17.0 & 44.9 & 24.6 & 15.5 \\
Sa2VA-4B~\citep{yuan2025sa2va} & 44.0 & 29.4 & 40.8 & 22.2 & 48.7 & 30.5 & 18.9 \\
Sa2VA-8B~\citep{yuan2025sa2va} & 43.2 & 27.5 & 38.3 & 20.4 & 49.3 & 28.8 & 17.9 \\
SAMTok-4B~\citep{zhou2026samtok} & 43.5 & 34.0 & 43.7 & 23.5 & 51.2 & 32.2 & 20.0 \\
SAMTok-8B~\citep{zhou2026samtok} & 44.6 & 33.4 & 43.4 & 23.0 & 51.5 & 31.8 & 19.7 \\
\midrule
\rowcolor{groupblue}
\multicolumn{8}{c}{\emph{Specialized grounding models trained on \ourdataset{}}} \\ \addlinespace[3pt]
GLaMM-7B~\citep{rasheed2024glamm} & 66.2 & 33.3 & 51.5 & 44.9 & 46.2 & 45.5 & 30.4 \\
Sa2VA-4B~\citep{yuan2025sa2va} & 68.6 & 39.5 & 54.4 & 49.2 & 54.3 & 51.6 & 34.5 \\
Sa2VA-8B~\citep{yuan2025sa2va} & 68.4 & 39.4 & 54.2 & 49.3 & 52.7 & 50.9 & 34.0 \\
SAMTok-4B~\citep{zhou2026samtok} & \underline{70.0} & 57.4 & 64.7 & \underline{66.6} & 64.1 & 65.3 & 44.1 \\
SAMTok-8B~\citep{zhou2026samtok} & \textbf{70.6} & \underline{57.6} & \underline{65.2} & \textbf{67.1} & \underline{64.5} & \underline{65.8} & \underline{44.5} \\
\midrule
\rowcolor{rowgray}
\ourmodel{}-4B (Ours) & 68.0 & \textbf{61.2} & \textbf{66.5} & 66.4 & \textbf{66.6} & \textbf{66.5} & \textbf{45.6} \\
\bottomrule
\end{tabular}
\end{table}

%% file: tables/sota_GCG.tex
\begin{table}[t]
\centering
\scriptsize
\tabcolsep=0.08cm
\setlength{\aboverulesep}{0pt}\setlength{\belowrulesep}{0pt}
\renewcommand{\arraystretch}{1.25}
\caption{\textbf{Results on grounded conversation generation (GCG).}
“(ft)” indicates models further finetuned on GCG after mixed training.
}
\label{tab:gcg}
\begin{NiceTabular}{lcccccw{c}{0em}ccccc}[colortbl-like]
\CodeBefore
  \rectanglecolor{groupblue}{2-2}{2-6}   %
  \rectanglecolor{groupblue}{2-8}{2-12}  %
\Body
\toprule
& \multicolumn{5}{c}{\textbf{Val}}
& & \multicolumn{5}{c}{\textbf{Test}} \\
Method              & METEOR & CIDEr & AP50 & mIoU & Recall & & METEOR & CIDEr & AP50 & mIoU & Recall \\ \midrule
Kosmos-2B~\citep{peng2024kosmos2} & 16.1 & 27.6 & 17.1 & 55.6 & 28.3 &  & 15.8 & 27.2 & 17.2 & 56.8 & 29.0 \\
BuboGPT-7B~\citep{zhao2023bubogpt} & \underline{17.2} & 3.6 & 19.1 & 54.0 & 29.4 &  & 17.1 & 3.5 & 17.3 & 54.1 & 27.0 \\
LISA-7B~\citep{lai2024lisa} & 13.0 & 33.9 & 25.2 & 62.0 & 36.3 &  & 12.9 & 32.2 & 24.8 & 61.7 & 35.5 \\
OMG-LLaVA-7B~\citep{zhang2024omgllava} & 14.9 & 41.2 & 29.9 & 65.5 & - &  & 14.5 & 38.5 & 28.6 & 64.7 & - \\
GLaMM-7B~\citep{rasheed2024glamm} & 16.2 & 47.2 & 30.8 & 66.3 & 41.8 &  & 15.8 & 43.5 & 29.2 & 65.6 & 40.8 \\
MGLMM-7B~\citep{zhou2024MGLMM} & 16.4 & 50.1 & 31.7 & 66.3 & 45.2 &  & - & - & - & - & - \\
Sa2VA-8B~\citep{yuan2025sa2va} & 16.4 & 49.5 & 33.2 & 67.7 & 45.1 &  & 16.2 & 49.0 & 32.2 & 66.8 & 44.5 \\
SAMTok-4B~\citep{zhou2026samtok} & 16.1 & 49.5 & 37.8 & 72.1 & 47.6 &  & 16.1 & 52.0 & 37.4 & 70.6 & 48.2 \\
SAMTok-7B~\citep{zhou2026samtok} & \underline{17.2} & \underline{54.8} & 38.2 & 72.6 & 48.9 &  & 17.0 & \underline{54.5} & 37.0 & 71.7 & 47.9 \\
SAMTok-7B (ft)~\citep{zhou2026samtok} & \textbf{17.7} & \textbf{55.0} & \underline{38.5} & \underline{72.9} & \underline{49.8} &  & \textbf{17.4} & 53.7 & 37.5 & 71.2 & 48.7 \\
\midrule
\rowcolor{rowgray}
\textbf{\ourmodel{}-4B} & 16.8 & 54.1 & 38.2 & 72.2 & 49.7 &  & 16.6 & 53.8 & \underline{38.7} & \underline{72.0} & \underline{50.9} \\
\rowcolor{rowgray}
\textbf{\ourmodel{}-4B (ft)} & \underline{17.2} & 54.4 & \textbf{39.6} & \textbf{73.7} & \textbf{50.9} &  & \underline{17.2} & \textbf{58.5} & \textbf{40.5} & \textbf{73.0} & \textbf{52.1} \\
\bottomrule
\end{NiceTabular}
\end{table}

%% file: tables/sota_refcoco.tex
\begin{table}[t]
\centering
\scriptsize
\tabcolsep=0.08cm
\setlength{\aboverulesep}{0pt}\setlength{\belowrulesep}{0pt}
\caption{\textbf{Results on referring expression segmentation (RES).} Performance is measured using cIoU. “(ft)” denotes models further finetuned on the corresponding RES benchmark after mixed training.
}
\label{tab:refcoco}
\renewcommand{\arraystretch}{1.25}

\begin{NiceTabular}{
    l
    ccc
    w{c}{0em}
    ccc
    w{c}{0em}
    cc
}[colortbl-like]
\CodeBefore
  \rectanglecolor{groupblue}{2-2}{2-4}    %
  \rectanglecolor{groupblue}{2-6}{2-8}    %
  \rectanglecolor{groupblue}{2-10}{2-11}  %
\Body
\toprule
& \multicolumn{3}{c}{\textbf{RefCOCO}}
& & \multicolumn{3}{c}{\textbf{RefCOCO+}}
& & \multicolumn{2}{c}{\textbf{RefCOCOg}} \\

Method
& Val & TestA & TestB
& & Val & TestA & TestB
& & Val (U) & Test (U) \\
\midrule
SAM 3 Agent (Gemini 2.5 Pro)~\citep{carion2025sam} & 75.5 & 77.6 & 71.0 &  & 67.3 & 71.1 & 63.4 &  & 73.4 & 74.0 \\
LISA-7B (ft)~\citep{lai2024lisa} & 74.9 & 79.1 & 72.3 &  & 65.1 & 70.8 & 58.1 &  & 67.9 & 70.6 \\
OMG-LLaVA-7B (ft)~\citep{zhang2024omgllava} & 78.0 & 80.3 & 74.1 &  & 69.1 & 73.1 & 63.0 &  & 72.9 & 72.9 \\
GLaMM-7B (ft)~\citep{rasheed2024glamm} & 79.5 & 83.2 & 76.9 &  & 72.6 & 78.7 & 64.6 &  & 74.2 & 74.9 \\
GSVA-13B (ft)~\citep{xia2024gsva} & 79.2 & 81.7 & 77.1 &  & 70.3 & 73.8 & 63.6 &  & 75.7 & 77.0 \\
PSALM-1.3B~\citep{zhang2024psalm} & 83.6 & 84.7 & 81.6 &  & 72.9 & 75.5 & 70.1 &  & 73.8 & 74.4 \\
EVF-SAM-1B~\citep{zhang2026evf} & 82.4 & 84.2 & 80.2 &  & 76.5 & 80.0 & 71.9 &  & 78.2 & 78.3 \\
SAM4MLLM-8B~\citep{chen2024sam4mllm} & 79.8 & 82.7 & 74.7 &  & 74.6 & 80.0 & 67.2 &  & 75.5 & 76.4 \\
SegLLM-7B~\citep{wang2025segllm} & 80.2 & 81.5 & 75.4 &  & 70.3 & 73.0 & 62.5 &  & 72.6 & 73.6 \\
Sa2VA-4B~\citep{yuan2025sa2va} & 81.7 & - & - &  & 77.4 & - & - &  & 80.0 & - \\
SAMTok-4B~\citep{zhou2026samtok} & 83.4 & \underline{85.0} & \underline{82.1} &  & \textbf{80.2} & \underline{83.4} & \underline{76.6} &  & 80.7 & 81.0 \\
\midrule
\rowcolor{rowgray}
\textbf{\ourmodel{}-4B} & \underline{83.8} & 84.8 & \textbf{82.7} &  & 79.6 & 83.0 & 75.7 &  & \underline{81.0} & \underline{81.8} \\
\rowcolor{rowgray}
\textbf{\ourmodel{}-4B (ft)} & \textbf{84.4} & \textbf{85.3} & \textbf{82.7} &  & \underline{79.8} & \textbf{83.7} & \textbf{77.2} &  & \textbf{81.1} & \textbf{82.2} \\
\bottomrule
\end{NiceTabular}
\end{table}

%% file: figures/examples.tex
\begin{figure}[!t]
\centering
\includegraphics[width=1\linewidth]{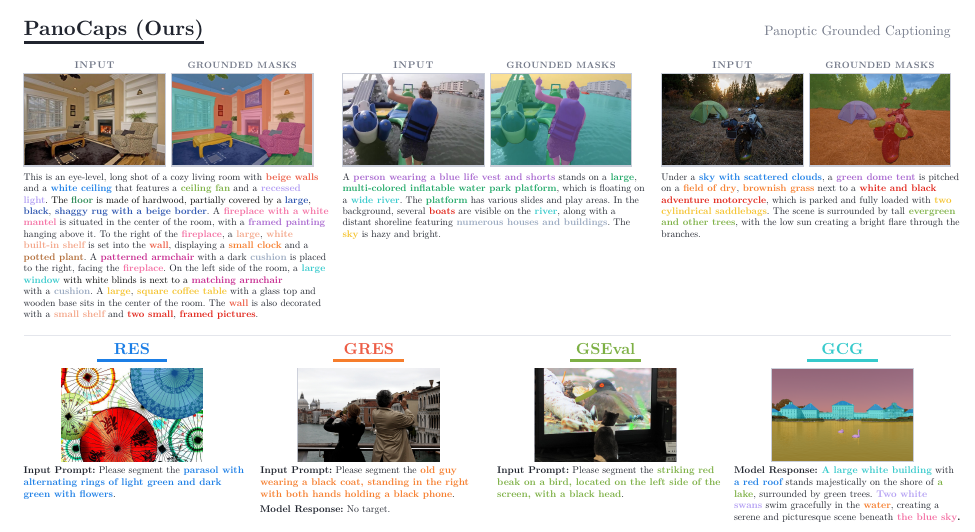}
\caption{\textbf{Qualitative results on grounding tasks.}
The top row shows \ourmodel{} on \ourdataset{}. The bottom row presents results on referring expression segmentation (RES), generalized referring expression segmentation (GRES), GroundingSuite (GSEval), and grounded conversation generation (GCG).
}
\label{fig:examples}
\end{figure}

%% file: tables/sota_GRES.tex
\begin{table}[t]
\centering
\scriptsize
\tabcolsep=0.08cm
\setlength{\aboverulesep}{0pt}\setlength{\belowrulesep}{0pt}
\renewcommand{\arraystretch}{1.25}
\caption{\textbf{Results on generalized referring segmentation (GRES)}.
“(ft)” indicates models further finetuned on GRES after mixed training. $^{\dagger}$ marks numbers reported by~\citet{xia2024gsva}.
}
\label{tab:gres}
\begin{NiceTabular}{
    l
    ccc
    w{c}{0em}
    ccc
    w{c}{0em}
    ccc
}[colortbl-like]
\CodeBefore
  \rectanglecolor{groupblue}{2-2}{2-4}    %
  \rectanglecolor{groupblue}{2-6}{2-8}    %
  \rectanglecolor{groupblue}{2-10}{2-12}  %
\Body
\toprule
& \multicolumn{3}{c}{\textbf{Val}}
& & \multicolumn{3}{c}{\textbf{TestA}}
& & \multicolumn{3}{c}{\textbf{TestB}} \\

Method
& gIoU & cIoU & N-acc
& & gIoU & cIoU & N-acc
& & gIoU & cIoU & N-acc \\
\midrule
LISA-7B$^{\dagger}$~\citep{lai2024lisa} & 32.2 & 38.7 & 2.7 &  & 48.5 & 52.6 & 6.4 &  & 39.7 & 44.8 & 5.0 \\
LISA-7B (ft)$^{\dagger}$~\citep{lai2024lisa} & 61.6 & 61.8 & 54.7 &  & 66.3 & 68.5 & 50.0 &  & 58.8 & 60.6 & 51.9 \\
SAM4MLLM-8B~\citep{chen2024sam4mllm} & 71.9 & 67.8 & 66.1 &  & 74.2 & 72.2 & 63.9 &  & 65.3 & 63.4 & 60.0 \\
HiMTok-8B~\citep{wang2025himtok} & 68.7 & 66.8 & - &  & 67.6 & 68.6 & - &  & 64.1 & 65.8 & - \\
HiMTok-8B (ft)~\citep{wang2025himtok} & 72.1 & 70.4 & - &  & 73.5 & 74.9 & - &  & 71.7 & \underline{72.0} & - \\
MLLMSeg-8B (ft)~\citep{wang2025mllmseg} & 75.1 & 71.6 & 73.2 &  & 77.0 & 76.9 & 72.4 &  & 69.7 & 68.5 & 65.5 \\
ARGenSeg-8B~\citep{wang2025argenseg} & 74.7 & 72.2 & - &  & 73.7 & 73.6 & - &  & 70.4 & 70.0 & - \\
SAMTok-4B~\citep{zhou2026samtok} & 74.7 & 71.0 & 68.8 &  & 76.9 & 75.8 & 66.2 &  & 69.1 & 68.3 & 60.8 \\
SAMTok-7B~\citep{zhou2026samtok} & 74.4 & 71.4 & 67.5 &  & 77.8 & 72.9 & 77.4 &  & 68.5 & 68.2 & 59.8 \\
SAMTok-7B (ft)~\citep{zhou2026samtok} & 78.0 & 72.1 & 79.5 &  & 78.5 & 77.3 & 75.9 &  & 69.6 & 67.7 & 68.6 \\
\midrule
\rowcolor{rowgray}
\textbf{\ourmodel{}-4B} & \underline{84.9} & \underline{75.1} & \underline{87.7} &  & \underline{79.8} & \underline{77.5} & \underline{85.5} &  & \underline{74.5} & 70.8 & \underline{81.2} \\
\rowcolor{rowgray}
\textbf{\ourmodel{}-4B (ft)} & \textbf{88.2} & \textbf{78.0} & \textbf{92.3} &  & \textbf{80.9} & \textbf{78.9} & \textbf{91.0} &  & \textbf{76.1} & \textbf{72.5} & \textbf{87.3} \\
\bottomrule
\end{NiceTabular}
\end{table}

%% file: tables/sota_groundingsuite.tex
\begin{table}[!t]
\centering
\scriptsize
\tabcolsep=0.08cm
\setlength{\aboverulesep}{0pt}\setlength{\belowrulesep}{0pt}
\renewcommand{\arraystretch}{1.25}
\caption{\textbf{Results on GroundingSuite (GSEval).} Performance is measured by gIoU, with all models evaluated zero-shot without training on GroundingSuite~\citep{hu2025groundingsuite}. $^{\dagger}$ denotes results reported by~\citet{hu2025groundingsuite}, and $^{\ddagger}$ denotes results reported by~\citet{yuan2026instructsam}.
}

\label{tab:grounding-suite}
\begin{tabular}{lccccc}
\toprule
 & \multicolumn{5}{c}{\textbf{GSEval}} \\
Method & \cellcolor{groupblue}Stuff & \cellcolor{groupblue}Part & \cellcolor{groupblue}Multi & \cellcolor{groupblue}Single & \cellcolor{groupblue}All \\
\midrule
LAVT$^{\dagger}$~\citep{yang2022lavt} & 6.0 & 10.7 & 45.0 & 25.8 & 22.5 \\
ReLA$^{\dagger}$~\citep{liu2023gres} & 3.8 & 8.8 & 46.7 & 19.7 & 19.7 \\
LISA-7B$^{\dagger}$~\citep{lai2024lisa} & 85.2 & 21.2 & 71.5 & 42.8 & 57.6 \\
GLaMM-7B$^{\dagger}$~\citep{rasheed2024glamm} & 86.9 & 16.5 & 70.4 & 42.1 & 57.2 \\
GSVA-7B$^{\dagger}$~\citep{xia2024gsva} & 76.0 & 20.0 & 57.8 & 34.2 & 48.6 \\
PSALM-1.3B$^{\dagger}$~\citep{zhang2024psalm} & 39.0 & 10.0 & 53.7 & 36.9 & 37.7 \\
EVF-SAM-1B$^{\dagger}$~\citep{zhang2026evf} & 85.1 & 23.1 & \underline{72.1} & 54.5 & 62.6 \\
InstructSeg-3B$^{\dagger}$~\citep{wei2025instructseg} & 56.2 & 24.2 & 66.8 & 51.3 & 52.5 \\
Sa2VA-4B$^{\ddagger}$~\citep{yuan2025sa2va} & 88.5 & 17.4 & 68.4 & 45.5 & 58.5 \\
Sa2VA-8B$^{\ddagger}$~\citep{yuan2025sa2va} & 77.3 & 18.0 & 72.0 & 45.5 & 56.3 \\
SAMTok-3B~\citep{zhou2026samtok} & \underline{90.2} & \underline{40.8} & 62.5 & \underline{63.6} & \underline{67.8} \\
\midrule
\rowcolor{rowgray}
\textbf{\ourmodel{}-4B} & \textbf{92.6} & \textbf{53.5} & \textbf{80.0} & \textbf{68.2} & \textbf{75.5} \\
\bottomrule
\end{tabular}
\end{table}

%% file: sec/08_conclusion.tex
\section{Conclusion}

To advance dense scene understanding, we study panoptic grounded captioning, a task that requires generating full-scene descriptions in which every phrase is grounded to pixel-level masks. We introduce \ourdataset{}, a benchmark of human-written captions with explicit phrase-mask alignments, combining free-form scene descriptions with near-complete pixel coverage and an evaluation protocol tailored to this setting.
Building on this benchmark, we propose \ourmodel{}, which formulates grounding as mask proposal selection. For each phrase, the VLM produces a concept vector that conditions a promptable segmenter to generate candidate masks, and a learned scorer selects the relevant subset, naturally supporting both singular and plural references.
Our results suggest that a contextual language representation can serve as an interface to a pretrained proposal space, allowing language generation and spatial delineation to leverage their respective strengths. Across panoptic grounded captioning and established grounding benchmarks, \ourmodel{} achieves strong performance and generalizes beyond dense scene description. Together, \ourdataset{} and \ourmodel{} provide data, evaluation, and a model for pairing dense language descriptions with pixel-level scene understanding.

%% file: sec/09_appendix.tex
\clearpage
\appendix           %
\section*{Appendix}

\section{\ourdataset{} Dataset}
\label{app:dataset}

This section provides further details on the design and construction of \ourdataset{}. We first describe the data collection and annotation pipeline, including the protocol used to produce human-written captions and phrase-level mask alignments. We then present additional dataset statistics and analyses that complement the overview in the main paper, highlighting the diversity, density, and quality of the resulting annotations.

\subsection{Data Sources} 
\input{figures/appendix/wordcloud}
\ourdataset{} is constructed by curating images and masks from established panoptic segmentation corpora: COCONut~\citep{deng2024coconut}, ADE20K~\citep{Zhou_2017_CVPR}, and VIPSeg~\citep{Miao_2022_CVPR}, following the protocol described in \hyperref[app:ann-protocol]{Appendix~\ref*{app:ann-protocol}}. These sources provide complementary visual domains and scene contexts.
COCONut, an extension of COCO, emphasizes everyday scenes containing common objects across indoor and outdoor environments. ADE20K contributes broad scene diversity and richly annotated indoor environments (e.g., kitchens, bedrooms, and living rooms) with extensive \emph{stuff} coverage. VIPSeg provides human-centric, video-derived frames capturing people interacting with objects and environments. From VIPSeg, we select one frame per video, typically the first frame. VIPSeg inherits the VSPW annotation pipeline~\citep{miao2021vspw}, in which masks are drawn manually at 1 frame per second and extended to 15 frames per second by label propagation; first frames therefore carry direct manual annotations rather than propagated ones.
\Cref{fig:app-examples} shows fully annotated training examples from each of the three sources, illustrating their complementary scene types.
To avoid cross-split contamination and ensure fair evaluation, \ourdataset{} draws training images only from the source training splits, and validation and test images only from the source evaluation splits. All images are drawn from these established, publicly released datasets and are used under their existing licenses. Summary statistics of the annotated portion are reported below. A word cloud illustrating the most frequent caption terms is shown in \Cref{fig:app-wordcloud}.

\subsection{Annotation Protocol}
\label{app:ann-protocol}

This section describes the multi-stage protocol used to construct the \ourdataset{} dataset.

\noindent \textbf{Image Selection (Stage 1).}
In Stage~1, we screen the source corpora and retain only images that (i) contain at least two segmented regions, (ii) exhibit clear and interpretable masks with minimal errors or fragmentation, (iii) achieve high mask coverage over the image, and (iv) improve dataset diversity (e.g., scene type, object density, indoor/outdoor settings, day/night conditions). Images failing any criterion are excluded. This stage ensures that masks are reliable, sufficiently cover the scene, and span a diverse range of visual conditions.

\noindent \textbf{Captioning and Grounding (Stage 2).}
In Stage~2, English-proficient annotators receive for each example: (i) the original RGB image, (ii) a panoptic-mask overlay where each segmented region is color-coded and labeled with its segment ID, and (iii) a label map linking each segment ID to its corresponding base class, which helps disambiguate visually similar or overlapping regions. Annotators compose a natural, detailed caption describing salient entities, attributes, relations, and activities, while explicitly linking each noun phrase to its corresponding mask ID. We follow a format similar to that used in COCONut-PanCap~\citep{deng2025coconutpancap}.

Captions are written in fluent prose, with entities grounded using tags of the form \texttt{$<$UID: object\_description$>$}, where \texttt{UID} corresponds to a mask ID. When a phrase refers to multiple instances, several IDs can be grouped (e.g., \texttt{$<$3,4:stacked cardboard boxes$>$}). At training time, each annotation tag is rewritten in place as \texttt{$<$p$>$ object\_description $<$/p$>$ [SEG]}, emitting one \texttt{[SEG]} per phrase whose target is the set of masks named by the tag's UIDs, with the surrounding caption text unchanged. Each tag contains a noun phrase only, while verbs remain in the sentence and articles appear outside the tag; for example, “A \texttt{$<$6:small brown dog$>$} lies …” or “the \texttt{$<$12:red sedan$>$} is parked beside a \texttt{$<$13:glass storefront$>$}.” Mask IDs may be reused when an entity is referenced multiple times. Annotators describe entities as specifically as possible based on visible evidence, favoring attributes such as color, material, size, or position (e.g., \texttt{$<$5:dark wooden table$>$} rather than \texttt{$<$5:table$>$}). Throughout the process, annotators describe only visible content and do not rely on automatic captioning tools.

To calibrate the annotation guidelines, we first conduct a pilot study on a small subset of images with our annotation partner. The annotation team is then trained according to the finalized protocol. Finally, all captions undergo an additional verification stage by a separate group of annotators to ensure compliance with the guidelines, correct phrase-mask alignment, and adequate scene coverage. This multi-stage process results in coherent captions with dense grounding and explicit links between masks and free-form noun phrases. Depending on the source dataset, each annotation batch involved up to 15 annotators and 8 reviewers. The median labeling time was $\sim$15 minutes per image and the median review time $\sim$12 minutes, amounting to $\sim$1{,}530 hours of human annotation effort in total. Annotators and reviewers were professionally contracted and compensated through our annotation partner, at a rate of \$1.35 per annotated image.

\input{tables/appendix/data_distribution}

\input{tables/appendix/dataset_stats}

\subsection{Additional Dataset Details}

The split-level statistics of \ourdataset{} are summarized in Table~\ref{tab:panocaps-stats} of the main paper, while the tables below provide a detailed breakdown of corpus composition, lexical statistics, and region characteristics for each split. The near-complete pixel coverage ($\approx 99\%$) and the average of nine grounded entities per image show that the annotations are dense rather than sparse, which is important for panoptic grounded captioning. The consistent coverage across splits indicates comparable annotation density, while the comparable number of masks per image (\Cref{fig:app-mask-percentage}) indicates similar grounding complexity across splits.

Table~\ref{tab:app-corpus-composition} reports the composition and split sizes for each source dataset. \ourdataset{} combines ADE20K~\citep{Zhou_2017_CVPR}, COCONut~\citep{deng2024coconut}, and VIPSeg~\citep{Miao_2022_CVPR}, while keeping source training and evaluation images in separate splits.

Table~\ref{tab:app-lexical-grounding} summarizes the language characteristics, including caption length and the number of unique free-form noun phrases (UNPs). The dataset exhibits a large variety of noun phrases, reflecting substantial lexical diversity beyond the source segmentation taxonomies.

Table~\ref{tab:app-region-stats} characterizes the region distribution. 
We report the proportion of \emph{things} versus \emph{stuff} masks and their area distribution (S/M/L). This mixture of categories and scales reflects the structural complexity of natural scenes, covering a range of spatial scales relevant to both grounding and caption generation.

Overall, these statistics show that \ourdataset{} combines dense pixel supervision, diverse scene composition, and lexically diverse free-form annotations. Additional qualitative examples are shown in \Cref{fig:app-examples}.

\subsection{Comparison with COCONut-PanCap}
\input{tables/appendix/comparison}
To better contextualize the properties of \ourdataset{}, we perform a direct comparison with the COCONut-PanCap dataset~\citep{deng2025coconutpancap}, which is the closest existing resource for panoptic grounded captioning. We provide additional details here.
To ensure a fair analysis, we identify a subset of 1,550 overlapping images that appear in both datasets. These images span both training and evaluation splits, allowing us to compare the annotation characteristics under the same visual content and segmentation masks.
Table~\ref{tab:app-pancap-compare} summarizes the key quantitative differences between the two datasets. Both datasets use the same underlying segmentation masks for the selected images; this controls for image content and segmentation masks, allowing the comparison to focus on differences in the captions and their grounding. Specifically, we report:

\begin{itemize}
\item \textbf{Masks referenced (\%)} --- the percentage of segmentation masks that are explicitly referenced in the caption. PanoCaps references a larger portion of the available masks, an increase of \textbf{26\%}.
\item \textbf{Unique labels} --- the number of distinct free-form phrases used to describe masks. PanoCaps exhibits \textbf{21\%} more unique labels, reflecting richer lexical diversity.
\item \textbf{Pixel coverage (\%)} --- the proportion of image pixels covered by the union of masks referenced in the caption. PanoCaps shows an increase of \textbf{50\%}, indicating that its descriptions capture a much larger portion of the visual scene.
\item \textbf{Masks referenced per image} --- the average number of masks mentioned in the caption. Importantly, we only count masks that are actually referenced in the caption, as these correspond to the entities that a grounded captioning model is expected to describe. Under this criterion, PanoCaps includes \textbf{23\%} more grounded masks per image.
\end{itemize}

\input{figures/appendix/pancap_caption_example}

\noindent \textbf{Cleaning the COCONut-PanCap Captions.}
The released COCONut-PanCap annotations contain noise, as illustrated in \Cref{fig:dataset}, and some captions cannot be reliably parsed into phrase-mask pairs. We therefore regenerate the grounded captions used in our training mixture with Qwen3-VL-235B~\citep{bai2025qwen3}. We prompt the model to describe each image while associating every mentioned entity with its corresponding panoptic segmentation mask, thereby linking each phrase to the appropriate mask ID. An LLM-based orchestrator then manages the verification and correction process, using Qwen3-VL and Stanza~\citep{qi2020stanza}, a dependency parser and lemmatizer, to identify and correct remaining issues. These include mismatches between the stated and actual number of grounded entities, repeated phrases between captions and entity tags, and malformed or invalid mask references. \Cref{fig:app-pancap-example} shows an example on a shared image. The cleaned caption provides more consistent phrase-level grounding, while the \ourdataset{} annotation remains denser, with more specific phrasing visible in this example.

Overall, on the same images and segmentation masks, \ourdataset{} references more of the available masks, attains higher pixel coverage, and uses a more varied vocabulary than COCONut-PanCap.

\section{\ourdataset{} Evaluation}

\label{app:panocaps_eval}

This section describes the prediction format and evaluation protocol used for panoptic grounded captioning on \ourdataset{}, including the open-text phrase-mask matching procedure and the metrics used to assess captioning, segmentation, and grounding quality. We then contrast this protocol with prior GCG evaluation and validate our phrase-similarity measure against human similarity judgments.

\subsection{Evaluation Protocol}

\noindent \textbf{Prediction Format.} Models are prompted to generate a grounded caption in which each phrase is associated with a segmentation mask. During training and inference, phrases are marked by \texttt{<p> object\_description </p>} tags followed by a single \texttt{[SEG]} token, whose hidden state provides the concept vector used for grounding (e.g., 
\textit{A <p> dog </p> \texttt{[SEG]} sits on <p> two pillows </p> \texttt{[SEG]}}). The \texttt{[SEG]} embeddings then condition the proposal model to generate candidate masks, from which the match scorer selects those corresponding to each phrase. This format differs slightly from the dataset annotation format, which uses tags of the form \texttt{<UID: object\_description>} to explicitly link noun phrases to ground-truth mask IDs. The prompts used during training and evaluation are shown in \Cref{app:panocaps-prompts}.

\noindent \textbf{Evaluation Metrics.}
We evaluate models on \ourdataset{} along three axes: caption quality, segmentation accuracy, and grounding quality. Caption quality is measured with CAPTURE~\citep{dong2024benchmarking}, while segmentation quality is evaluated using AP50 and mean IoU (mIoU). Grounding performance is evaluated using the phrase-mask matching procedure described in Procedure~\ref{alg:matching}.

Each predicted or ground-truth mask is associated with a set of phrases, and text similarity between two masks is defined as the maximum similarity over their associated phrase pairs. Exact matches, after lowercasing and removal of a leading article, receive a score of $1.0$. For single-token phrases we additionally assign $1.0$ when the two phrases share a WordNet synset, comparing lemmatized forms~\citep{miller1995wordnet}. Otherwise, Sentence-BERT~\citep{reimers2019sentence} provides a similarity score in $[0,1]$, computed with all-mpnet-base-v2, the highest-quality pretrained model of the Sentence-Transformers library. We refer to this combination of exact, lexical, and embedding-based matching as our phrase-similarity measure. Predicted masks with $\mathrm{IoU}\ge0.9$ and text similarity $\ge0.5$ are merged before matching.

We perform one-to-one Hungarian matching, using the mean of the text and mask similarities as the assignment score. A matched pair is considered correct when both similarities are at least $0.5$. We report Precision, Recall, and F1 based on the resulting correct matches:
\begin{equation}
\mathrm{Precision}
= \frac{\#\text{correct matches}}{\#\text{predictions}} ,
\end{equation}

\begin{equation}
\mathrm{Recall}
= \frac{\#\text{correct matches}}{\#\text{ground-truth masks}}.
\end{equation}

We additionally report a generalized Panoptic Quality (gPQ) that uses the graded similarities, as defined in \Cref{eqn:gpq} of the main paper. Unlike the binary matching metrics, gPQ preserves the graded quality of both spatial and textual agreement.

\input{figures/appendix/metric_cases}

\input{figures/appendix/annotated_examples}

\input{figures/appendix/prompts_panocaps}

\subsection{Comparison with Prior GCG Evaluation}

Our procedure differs from GLaMM's protocol~\citep{rasheed2024glamm} in three respects: the matching procedure, the text-similarity function, and the reported metrics. First, prior GCG evaluation matches predictions to ground truth greedily by mask IoU, gated by a soft text-similarity check, whereas we perform a global assignment on the mean of textual and spatial similarity, resolving ambiguous correspondences jointly rather than in discovery order. Second, the two protocols differ in how phrase similarity is scored, which we validate below. Third, prior work reports mIoU and Recall, which are computed over matched pairs only and therefore do not penalize spurious predictions. We additionally report Precision, which exposes spurious groundings, and gPQ, which combines the quality of matched pairs with their F1 and thus reflects missed and spurious regions in a single panoptic measure.

\noindent \textbf{Validation of the Text-Similarity Metric.}
To validate our phrase-similarity measure, we compare it with the scorer used in prior GCG evaluation, which computes cosine similarity between mean-pooled BERT embeddings~\citep{devlin2019bert} and accepts pairs above a threshold of $0.5$. We evaluate both measures on BiRD~\citep{asaadi2019big}, a dataset of 3,345 bigram pairs annotated with fine-grained human relatedness scores, providing a proxy for the short noun phrases encountered in our evaluation. Our measure correlates substantially better with human judgments (Spearman $\rho=0.72$ vs.\ $\rho=0.45$). The gap arises because BERT cosine similarities tend to cluster in a narrow, high-valued range: $98.8\%$ of BiRD pairs, including unrelated ones, exceed the $0.5$ threshold, and pairs in the lowest human-relatedness quartile still obtain an average similarity of $0.68$. In contrast, our measure assigns an average similarity of $0.26$ to the same pairs, compared with $0.67$ for the highest quartile, making the $0.5$ threshold a meaningful operating point.

\Cref{fig:app-metric-cases} illustrates both differences on a \ourdataset{} validation image. In example (a), a prediction grounds a plant pot with an accurate mask but an unrelated phrase; the pair passes the BERT gate and is counted as correct under the prior protocol, while our metric rejects it. In example (b), two hallucinated predictions appear alongside otherwise correct groundings; the prior protocol's matched-pair metrics remain perfect, whereas Precision and gPQ expose them.

\section{Evaluation of Generalist VLMs on \ourdataset{}}
\label{app:generalist-prompt}

We additionally evaluate two generalist vision-language models on \ourdataset{} using the evaluation protocol described above.
Table~\ref{tab:panocaps} compares \ourmodel{} with both models on \ourdataset{}: Gemini 2.5 Pro~\citep{comanici2025gemini}, a closed-source commercial API, and Qwen3-VL-235B-A22B~\citep{bai2025qwen3}, an open-source model.
In this section, we describe how these models are prompted and evaluated on \ourdataset{}.

Since neither model natively supports generating a full caption and pixel-level masks in a single output, we adopt a two-stage protocol to evaluate them on \ourdataset{}. In the first stage, we prompt the models to produce an image caption with interleaved entity tags. Entity tags follow the format \texttt{$<$UID: object\_description$>$}, where \texttt{UID} is a unique identifier for each entity. For multiple instances of the same entity type, we allow grouped tags such as \texttt{<4,5,6: three birds>}, matching our annotation style. The model outputs a JSON dictionary with the caption stored under the key \texttt{"caption"}. We parse the JSON output to extract the set of mentioned entities, which we refer to as the \emph{entity list}. The captioning prompts used for Gemini 2.5 Pro and Qwen3-VL-235B-A22B are shown in Figures~\ref{app:gemini-prompts} and~\ref{app:qwen-prompts}, respectively.

In the second stage, given the image and the parsed entity list, we prompt the model to predict a spatial grounding for each entity. For Gemini 2.5 Pro, the segmentation prompt requests a JSON list containing 2D bounding boxes, segmentation masks, and the corresponding text labels. For Qwen3-VL-235B-A22B, which only supports box-level grounding, we instead request a JSON list of bounding boxes and labels and then derive segmentation masks by running SAM~\citep{ravi2024sam} on the predicted boxes using the default configuration. In both cases, labels are required to exactly match the input tags (including the \texttt{UID}) so that we can reliably align predicted masks with caption entities. 

We keep the overall prompt structure as similar as possible across models but introduce small model-specific adjustments to address failure modes. For example, Gemini 2.5 Pro follows the requested format without examples, whereas Qwen3-VL-235B-A22B benefits from few-shot examples that improve instruction following and JSON formatting. Finally, for robustness, if a model call returns an error or a non-parsable response, we retry the request up to five times; if all attempts fail, we record no prediction for that image. This two-stage protocol allows us to evaluate generalist VLMs on \ourdataset{} despite their limited support for joint caption-mask generation.

\input{figures/appendix/prompts_gemini}

\input{figures/appendix/prompts_qwen}

\section{Additional Results}
\label{app:additional-results}

\subsection{Trainable Set of the Proposal Model}
\label{app:ablation-trainable}

\Cref{tab:ablation-trainable} completes the component ablations of \Cref{sec:ablations} by varying which parameters of the proposal model are trained. Keeping the proposal model entirely frozen matches the default configuration, which unfreezes only the fusion encoder and match scorer, on gPQ, precision and AP50, with a small recall gain ($+0.8$) and a 0.4-point RefCOCO reduction. Additionally unfreezing the DETR decoder and mask heads yields no measurable gain on any metric while increasing training time by roughly 10\%. Together these results indicate that adapting the proposal model further is unnecessary once the interface to it is learned.

\input{tables/appendix/ablation_trainable}

\subsection{Selection Rule}
\label{app:selection-rule}

\Cref{tab:app-selection-rule} varies the inference-time selection rule. 
Since $\theta$ and the rule are applied only at inference, every arm reuses the same three trained checkpoints. On the left, performance peaks at the default value of 0.5, and the ordering is identical in all three seeds, so 0.5 is the best setting on gPQ in every run even where the difference from a neighboring value is not individually significant. On the right, replacing thresholding with a fixed number of proposals per phrase degrades sharply: keeping one collapses phrases that refer to several instances, while keeping two leaves recall unchanged and reduces precision by 26.9 points, converting singular referents into false positives. Captioning performance is unaffected, since the selection rule does not change the caption.

\input{tables/appendix/ablation_selection_rule}

\subsection{Efficiency}
\label{app:efficiency}

Table~\ref{tab:app-trainable} reports the total and trainable parameter counts, together with single GPU inference cost, for every finetuned model compared in Table~\ref{tab:panocaps}. We measure inference cost on the \ourdataset{} test split with batch size 1 on a single A100 GPU, reporting the median latency over 100 images for the full grounded captioning pipeline. Baseline trainable parameter sets follow each method's released training recipe. For \ourmodel{}, we train LoRA adapters, the token embeddings and output head, the concept bridge, and the proposal model's fusion encoder and match scorer.
Of SAM 3’s 859.9M parameters, we retain only the 486.7M-parameter image detector, discarding the 353.7M CLIP text tower and 19.6M video tracker.

\input{tables/appendix/trainable_params}

%% file: figures/appendix/wordcloud.tex
\begin{wrapfigure}{r}{0.5\linewidth}
\centering
  \includegraphics[width=\linewidth]{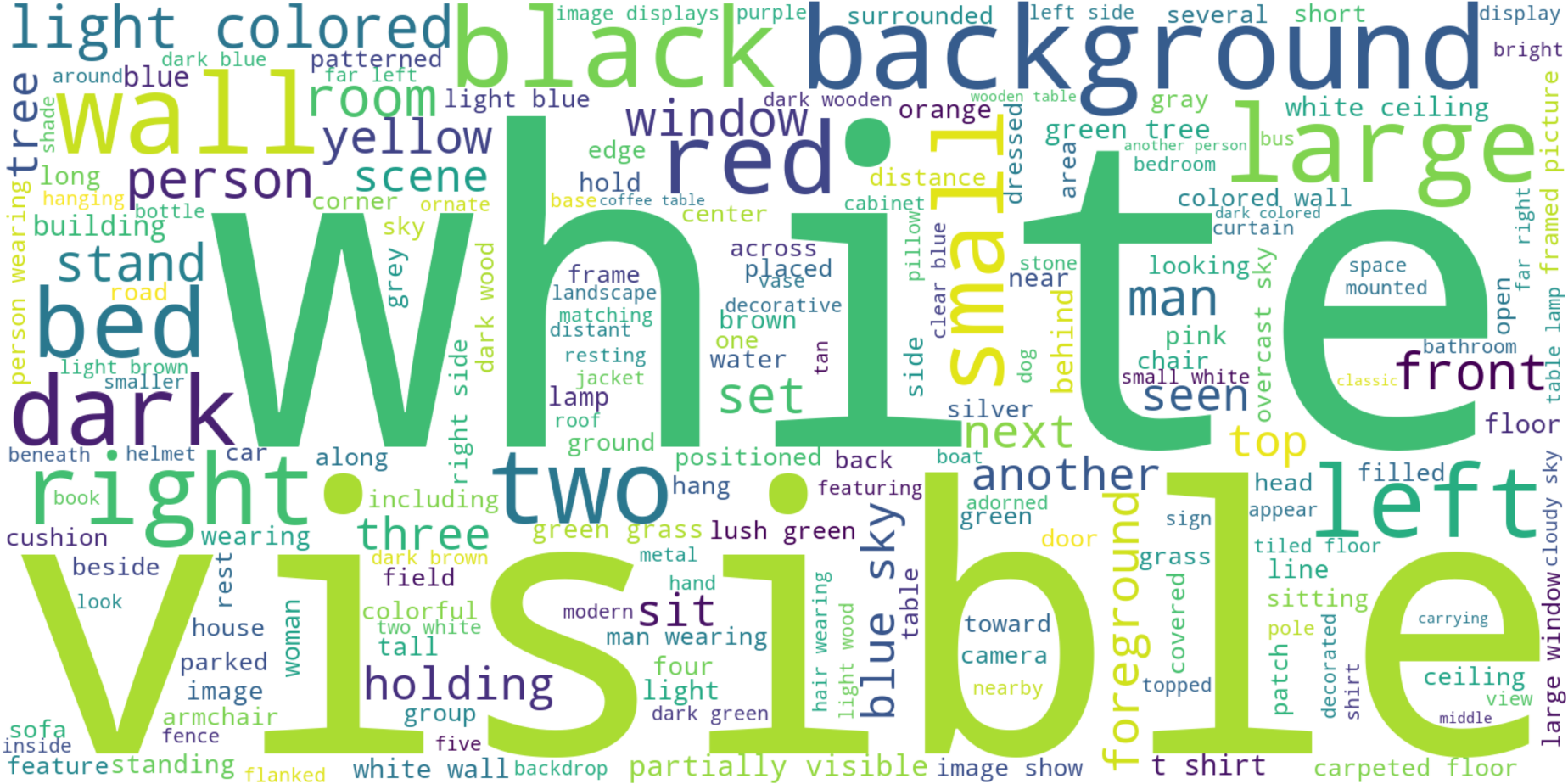}
    \caption{\textbf{Word cloud of caption terms in \ourdataset{}.}}
  \label{fig:app-wordcloud}
\end{wrapfigure}

%% file: tables/appendix/data_distribution.tex
\begin{table}[t]
    \centering
    \setlength{\aboverulesep}{0pt}\setlength{\belowrulesep}{0pt}
    \renewcommand{\arraystretch}{1.25}
    \caption{\textbf{Data distribution.} Image sources and per-split image counts.}
    \scriptsize
    \setlength{\tabcolsep}{4pt}
    \begin{tabular}{lcccc}
    \toprule
    \rowcolor{groupblue}
    Data &  Image Source & Train & Test &  Val  \\ \midrule
    ADE20K~\citep{Zhou_2017_CVPR}  & SUN~\citep{xiao2010sun}, Places~\citep{zhou2017places} & 570 & 245 &  105  \\
    COCONut~\citep{deng2024coconut} & COCO~\citep{lin2014microsoft} & 750 & 560 &  240   \\
    VIPSeg~\citep{Miao_2022_CVPR}  & VSPW~\citep{miao2021vspw} & 750 & 175 &  75  \\
    \bottomrule
    \end{tabular}
    \label{tab:app-corpus-composition}
\end{table}

%% file: tables/appendix/dataset_stats.tex
\begin{table*}[t]
\centering
    \begin{minipage}[t]{0.46\textwidth}
        \centering
        \input{tables/appendix/lexical_diversity}
    \end{minipage}
\hfill
    \begin{minipage}[t]{0.46\textwidth}
        \centering
        \input{tables/appendix/region_characteristics}
    \end{minipage}
\end{table*}

%% file: tables/appendix/lexical_diversity.tex
\setlength{\aboverulesep}{0pt}\setlength{\belowrulesep}{0pt}
\renewcommand{\arraystretch}{1.25}
\caption{\textbf{Lexical diversity.} Split-wise caption length and vocabulary richness. \emph{UNP} is the number of distinct noun phrases across the splits (higher is more diverse).}
\scriptsize
\setlength{\tabcolsep}{2pt}
\begin{tabular}{lccccc}
\toprule
\rowcolor{groupblue}
Split & \# Tokens & Words/cap & Tokens/cap & \# Entities &  UNP   \\ \midrule
Train & 196.6K & 80.5  & 95.0  & 19.8K   & 12.3K \\
Test  & 81.0K  & 69.7  & 82.7  & 7.9K    & 5.4K  \\
Val   & 35.4K  & 71.1  & 84.3  & 3.5K    & 2.7K  \\ \midrule
\rowcolor{rowgray}
All   & 313K & 76.3  & 90.2  & 31.3K   & 17.9K \\
\bottomrule
\end{tabular}
\label{tab:app-lexical-grounding}

%% file: tables/appendix/region_characteristics.tex
\caption{\textbf{Region characteristics.} Per-split mask size and type distribution.
\emph{Mask (S/M/L)} reports the share of masks with area $<$1\%, 1--10\%, and $>$10\% of the image.}
\label{tab:app-region-stats}
\setlength{\tabcolsep}{2pt}
\setlength{\aboverulesep}{0pt}\setlength{\belowrulesep}{0pt}
\renewcommand{\arraystretch}{1.25}
\scriptsize
\begin{tabular}{lcccc}
\toprule
\rowcolor{groupblue}
Split & Mask W$\times$H & Mask (S/M/L)  & Things (\%) & Stuff (\%) \\
\midrule
Train & $\num{323}\times\num{198}\,$ & 32/41/28 & 63.8     & 36.2    \\ %
Test  & $\num{304}\times\num{192}\,$ & 29/39/32 & 62.1     & 37.9    \\ %
Val   & $\num{305}\times\num{190}\,$ & 30/39/32 & 62.2     & 37.8    \\ \midrule %
\rowcolor{rowgray}
All   & $\num{316}\times\num{196}\,$ & 31/40/29 & 63.2     & 36.8    \\ %
\bottomrule
\end{tabular}

%% file: tables/appendix/comparison.tex
\begin{table*}[t]
\centering
    \begin{minipage}[t]{0.44\textwidth}
        \vspace{0pt}
        \centering
        \input{figures/appendix/mask_count}
    \end{minipage}
\hfill
    \begin{minipage}[t]{0.53\textwidth}
        \centering
        \input{tables/appendix/pancap_comparison}
    \end{minipage}
\end{table*}

%% file: figures/appendix/mask_count.tex
\includegraphics[width=\linewidth]{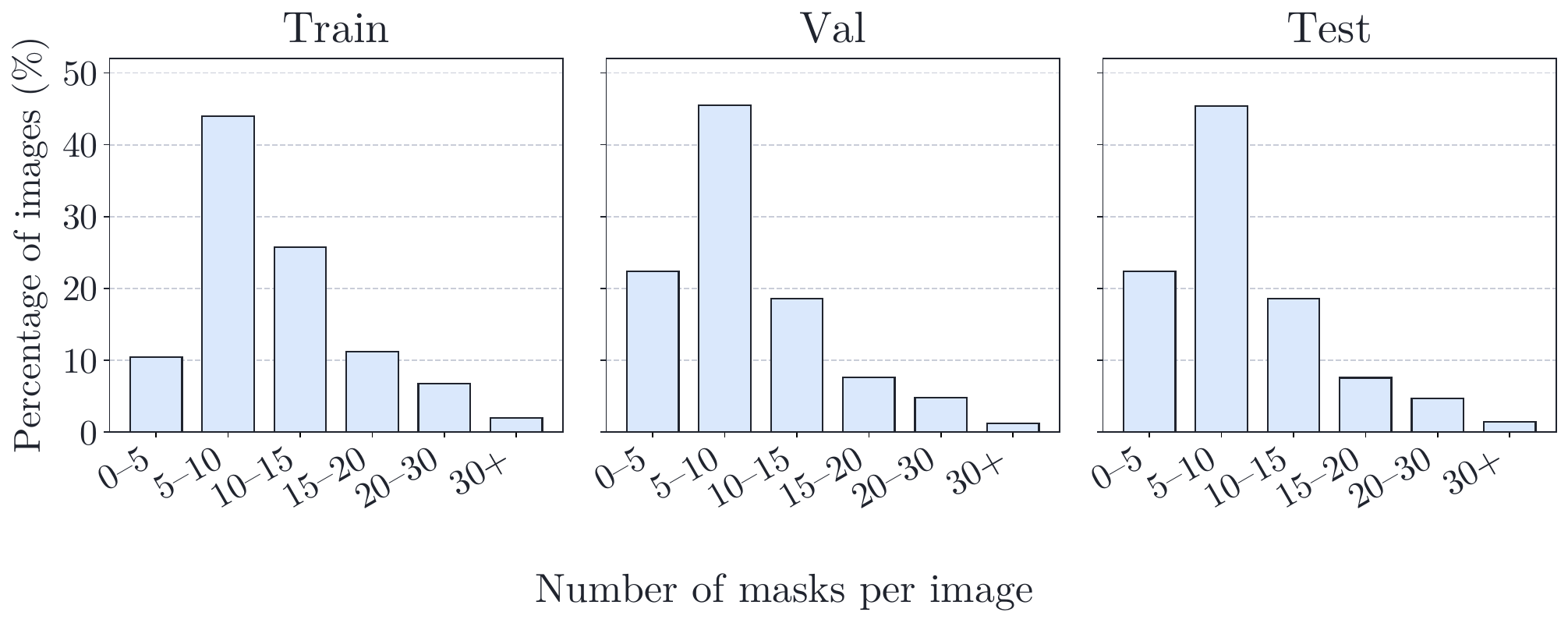}
\captionof{figure}{\textbf{Distribution of mask counts per image.} Each panel shows the percentage of images falling into bins of mask count, illustrating that all splits share a similar object density.}
\label{fig:app-mask-percentage}

%% file: tables/appendix/pancap_comparison.tex
\setlength{\aboverulesep}{0pt}\setlength{\belowrulesep}{0pt}
\renewcommand{\arraystretch}{1.25}
\caption{\textbf{Comparison of annotation density and vocabulary diversity} between COCONut-PanCap and \textbf{PanoCaps}, computed on the 1{,}550 images shared by both datasets. Relative improvements of PanoCaps over COCONut-PanCap are shown in gray.}
\label{tab:app-pancap-compare}
\scriptsize
\setlength{\tabcolsep}{3pt}
\begin{tabular}{lcc}
\toprule
\rowcolor{groupblue}
Metric & COCONut-PanCap & PanoCaps (Ours) \\
\midrule
Masks referenced (\%) & 79.2 & 99.9 {\color{gray}(+26\%)} \\
Unique labels (K) & 6.3 & 7.6 {\color{gray}(+21\%)} \\
Grounded phrases & 6349 & 7553 {\color{gray}(+19\%)} \\
Pixel coverage (\%) & 65.4 & 98.2 {\color{gray}(+50\%)} \\
Masks per image & 6.2 & 7.6 {\color{gray}(+23\%)} \\
\bottomrule
\end{tabular}

%% file: figures/appendix/pancap_caption_example.tex
\begin{figure*}[ht]
  \centering
  \includegraphics[width=\linewidth]{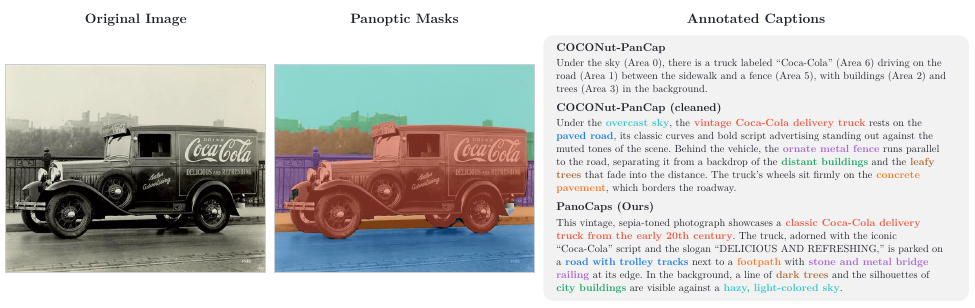}
  \caption{\textbf{Caption comparison on a shared image.} The same COCO image with its panoptic masks (left) and three annotations. The original COCONut-PanCap caption marks regions with bare (Area $n$) tags that do not delimit the referring phrase, so phrase-mask pairs cannot be reliably parsed (shown uncolored). The cleaned counterpart used in our training mixture fixes the format. The \ourdataset{} caption grounds more regions and adds specific, visually verifiable detail (e.g., the truck's slogan, its era, the trolley tracks), illustrating the quality gap between automatic and human annotation.
  }
  \label{fig:app-pancap-example}
\end{figure*}

%% file: figures/appendix/metric_cases.tex
\begin{figure*}[t]
  \centering
  \includegraphics[width=\linewidth]{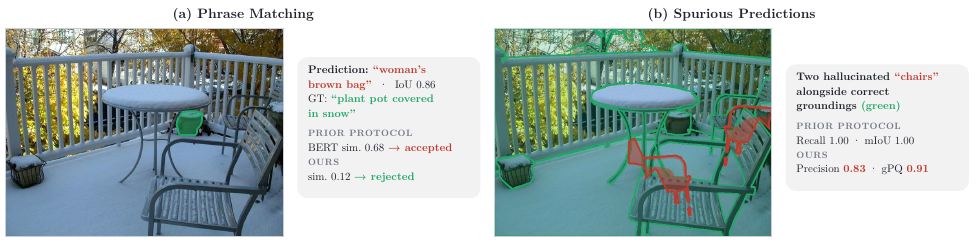}
\caption{\textbf{Failure modes of the prior GCG evaluation protocol.} Illustrated on a \ourdataset{} validation image. (a) A prediction grounds the ground-truth plant pot almost perfectly but calls it ``woman's brown bag'' instead of ``plant pot covered in snow''; the prior BERT gate scores the pair 0.68 and accepts the match, whereas our phrase-similarity measure scores it 0.12 and rejects it. (b) Two hallucinated ``chairs'' are predicted alongside otherwise correct groundings (green). Recall and mIoU, computed over matched pairs only, remain perfect, while Precision and gPQ expose the spurious masks.
}
  \label{fig:app-metric-cases}
\end{figure*}

%% file: figures/appendix/annotated_examples.tex
\begin{figure*}[th] %
  \centering
  \includegraphics[width=\linewidth]{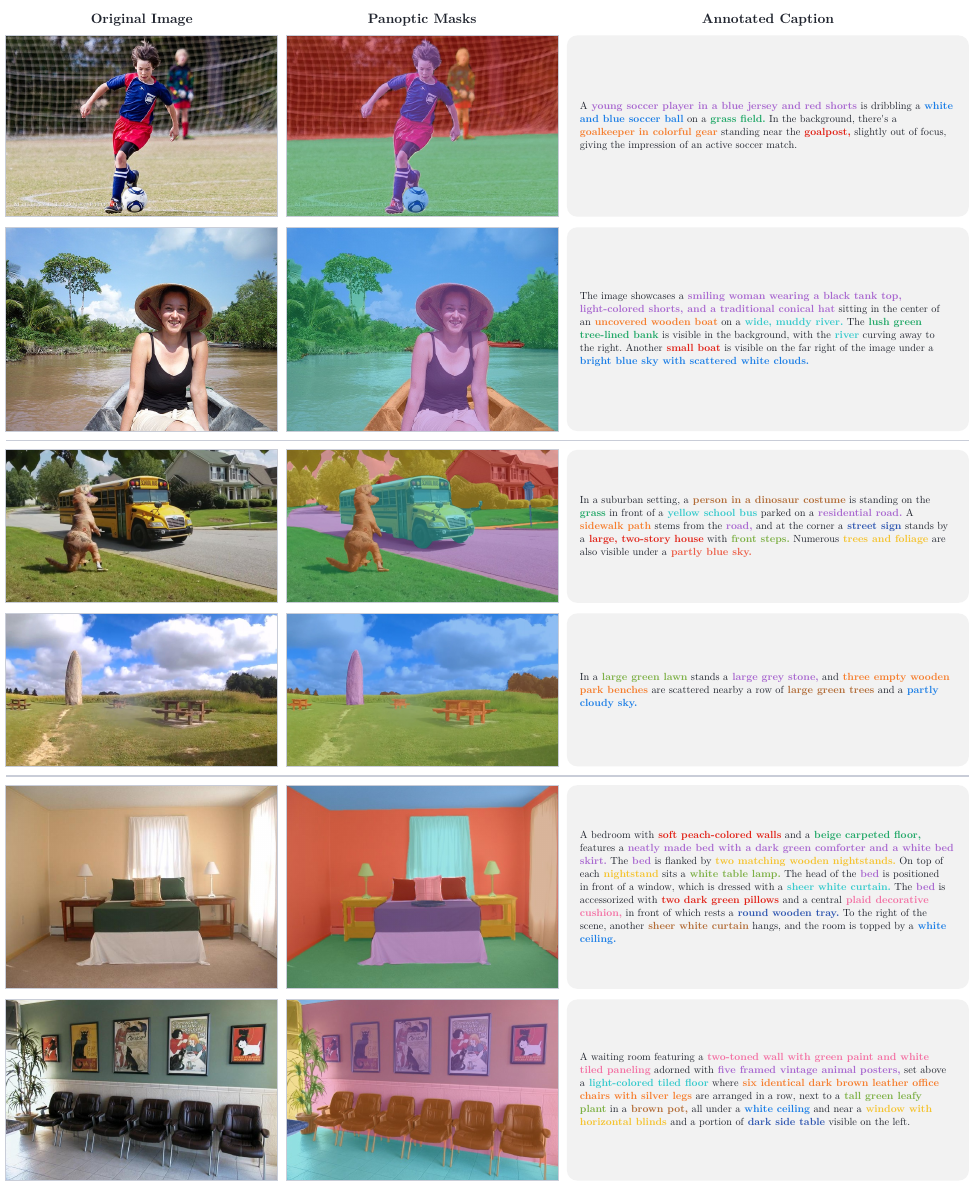}
  \caption{\textbf{\ourdataset{} annotated examples.} Training samples shown as the original image, the panoptic mask overlay, and the grounded caption, with every phrase colored to match its mask; phrases grounded to several masks at once, such as the three benches, share a single color. Horizontal rules separate the three image sources: COCONut (top two), VIPSeg (middle two), and ADE20K (bottom two).}
  \label{fig:app-examples}
\end{figure*}

%% file: figures/appendix/prompts_panocaps.tex
\begin{figure}[t]
\centering
\finding{5}{
\textbf{Task Input Prompts:}\\
Caption this image exhaustively: mention every object and every background region, and insert its segmentation mask right after each mention.\\
Write a complete scene caption that leaves nothing out --- every foreground object and background area must appear, each followed by its segmentation mask.\\
Write a caption covering all objects and background regions in the image, inserting the matching segmentation mask after each phrase.\\
Produce a panoptic caption of the scene: name all things and background regions, placing the segmentation mask immediately after each named element.\\
Give a full-scene caption in which every visible object and background area is mentioned, attaching the segmentation mask after each one.\\
Compose an exhaustive caption naming every region in the image, foreground and background alike, with the correct segmentation mask after each element you mention.\\
}
\caption{\textbf{Prompt templates used for training and evaluating models on \ourdataset{}.}}
\label{app:panocaps-prompts}
\end{figure}

%% file: figures/appendix/prompts_gemini.tex
\begin{figure}[t]
\centering
\begin{subfigure}[t]{0.95\linewidth}
\centering
\finding{3}{
\noindent\textbf{General Task Description:}\\
Give a detailed description of the scene in the image by describing \textbf{all} objects and any events or actions happening.\\
\textbf{Format Definition:}\\
Tag objects within $<>$ tags.\\
Do not tag the whole scene/room with $<>$ tags.\\
Do not tag parts of objects with $<>$ tags, but only the whole object.\\Include attributes of objects within the $<>$ tags resulting in unique and detailed object descriptions.\\
A tag should be of the format $<$UID: object\_description$>$, where UID is a unique identifier of each object, e.g. $<$1: red car$>$.\\
Do not include articles within $<>$.\\
If object\_description contains a numerical describing multiple objects of same type then include a list of UIDs, e.g. $<$4,5,6: three birds$>$.\\
\textbf{Output Structure:}\\Output a JSON dictionary of the caption in the key ``caption''.
}
\caption{Captioning prompt.}
\end{subfigure}
\vspace{1em}
\begin{subfigure}[t]{0.95\linewidth}
\centering
\finding{4}{
\textbf{General Task Description:}\\
Give the segmentation mask for \{object\_list\}.\\
\textbf{Format Definition:}\\
The numerical before each object description is a unique identifier for each object.\\
When there are multiple identical object descriptions with different unique identifiers, segment all instances.\\
\textbf{Output Structure:}\\
Output a JSON list of segmentation masks where each entry contains the 2D bounding box in the key ``box\_2d'', the segmentation mask in key ``mask'', and the text label in the key ``label''.\\
The labels should be identical to the input labels, including the unique identifier.
}
\caption{Segmentation prompt.}
\end{subfigure}
\caption{\textbf{Prompts used for caption generation and segmentation with Gemini 2.5 Pro.}}
\label{app:gemini-prompts}
\end{figure}

%% file: figures/appendix/prompts_qwen.tex
\begin{figure}[t]
\centering
\begin{subfigure}[t]{0.95\linewidth}
\centering
\finding{5}{
\textbf{General Task Description:}\\
 Give a detailed description of the scene in the image by describing \textbf{all} objects.\\
\textbf{Format Definition:}\\
Tag objects within $<>$ tags.\\
A tag should be of the format $<$UID:object\_description$>$, where UID is a unique identifier of each object, e.g. $<$1: red car$>$.\\
Include attributes of objects within the $<>$ tags resulting in unique and long detailed object descriptions.\\
Do not include articles within $<>$.\\
If there are multiple objects of the same object type then include a list of UIDs, e.g. $<$4,5,6: three birds$>$.\\
CAUTION: Object descriptions within $<>$ tags will be used for object detection, so ensure that you \textbf{ALWAYS} tag objects that you can see in the image.\\
\textbf{Output Structure:}\\    
Output a JSON dictionary of the caption in the key ``caption''.\\
Example 1 (BAD: objects not tagged):\\
\{``caption'': ``A red car parked on the street next to a tree and a dog.''\}\\
Example 2 (GOOD: every object tagged):\\
\{``caption'': ``A $<$1: red car$>$ parked on street next to $<$2: tall green tree$>$ and $<$3: small white dog$>$.''\}\\
Example 3 (GOOD: multiple similar objects):\\
\{``caption'': ``$<$1,2,3: three identical blue chairs$>$ arranged around $<$4: round wooden table$>$ in center of room.''\}
}
\caption{Captioning prompt.}
\end{subfigure}
\vspace{1em}
\begin{subfigure}[t]{0.95\linewidth}
\centering
\finding{6}{
\textbf{General Task Description:}\\
Detect the following objects in the image: \{object\_list\}. \\
\textbf{Format Definition:}\\
The numerical before each object description is a unique identifier for each object.\\
When there are multiple identical object descriptions with different unique identifiers, detect all instances.\\
\textbf{Output Structure:}\\  
Output a JSON list of bounding boxes where each entry contains the 2D bounding box in the key ``box\_2d'', and the text label in the key ``label''. The labels should be identical to the input labels including the unique identifier.\\
Bounding boxes should be in the format [x1,y1,x2,y2].
}
\caption{Detection prompt.}
\end{subfigure}
\caption{\textbf{Prompts for caption generation and object detection with Qwen3-VL-235B-A22B.}}
\label{app:qwen-prompts}
\end{figure}

%% file: tables/appendix/ablation_trainable.tex
\providecommand{\sd}[1]{{\tiny$\pm$#1}}
\providecommand{\dt}[1]{\,{\color{gray}\scriptsize(#1)}}
\providecommand{\dtb}[1]{\,{\scriptsize\textbf{(#1)}}}
\begin{table}[t]
\centering
\scriptsize
\tabcolsep=0.15cm
\setlength{\aboverulesep}{0pt}\setlength{\belowrulesep}{0pt}
\renewcommand{\arraystretch}{1.25}
\caption{\textbf{Trainable set of the proposal model (\ourmodel{}-2B).} Each variant is evaluated on the \ourdataset{} validation split and modifies a single component of the default model while keeping the training data, schedule, and effective batch size fixed. RefCOCO reports mean cIoU over the three validation splits. Every configuration is reported as the mean of three training seeds and $\pm$ is the standard deviation across them. Gray values denote changes relative to the default; \textbf{bold} marks statistically significant differences from the default ($p<0.05$, three seeds); the same convention is used in \Cref{tab:ablation-selection} and \Cref{tab:ablation-concept}.
}
\resizebox{\linewidth}{!}{%
\label{tab:ablation-trainable}
\begin{tabular}{lcccccc}
\toprule
\textbf{Configuration} & gPQ & Recall & Precision & AP50 & CAPTURE & RefCOCO \\
\midrule
\rowcolor{groupblue}
\ourmodel{}-2B (default) & 42.1\sd{0.6} & 60.3\sd{0.5} & 63.9\sd{1.2} & 60.8\sd{1.0} & 66.2\sd{0.2} & 80.4\sd{0.1} \\
\midrule
\quad frozen proposal model & 42.2\sd{0.7}\dt{+0.1} & 61.1\sd{0.6}\dt{+0.8} & 63.9\sd{0.9}\dt{0.0} & 60.8\sd{0.3}\dt{0.0} & 66.2\sd{0.3}\dt{0.0} & 80.0\sd{0.2}\dtb{-0.4} \\
\quad unfreeze decoder \& mask heads & 42.1\sd{0.3}\dt{0.0} & 60.6\sd{0.8}\dt{+0.3} & 64.3\sd{0.5}\dt{+0.4} & 59.8\sd{1.7}\dt{-1.0} & 66.4\sd{0.3}\dt{+0.2} & 80.1\sd{0.1}\dtb{-0.3} \\
\bottomrule
\end{tabular}
}
\end{table}

%% file: tables/appendix/ablation_selection_rule.tex
\providecommand{\sd}[1]{{\tiny$\pm$#1}}
\providecommand{\dt}[1]{\,{\color{gray}\scriptsize(#1)}}
\providecommand{\dtb}[1]{\,{\scriptsize\textbf{(#1)}}}
\begin{table}[t]
\centering
\scriptsize
\tabcolsep=0.10cm
\setlength{\aboverulesep}{0pt}\setlength{\belowrulesep}{0pt}
\renewcommand{\arraystretch}{1.25}
\caption{\textbf{Selection rule for \ourmodel{}-2B on the \ourdataset{} validation split.} Each configuration is the mean of three training seeds and $\pm$ is the standard deviation across them. Gray values denote changes relative to the default; \textbf{bold} marks statistically significant differences ($p<0.05$, three seeds).
}
\label{tab:app-selection-rule}
\resizebox{\linewidth}{!}{%
\begin{tabular}{lcccc@{\hspace{12pt}}lcccc}
\toprule
\multicolumn{5}{c}{\textit{Threshold sensitivity}} & \multicolumn{5}{c}{\textit{Fixed cardinality}} \\
\cmidrule(lr){1-5}\cmidrule(lr){6-10}
\rowcolor{groupblue}
\textbf{Threshold} & gPQ & Recall & Prec. & AP50 & \textbf{Proposals kept} & gPQ & Recall & Prec. & AP50 \\
\midrule
$\theta = 0.5$ (default) & 42.1\sd{0.6} & 60.3\sd{0.5} & 63.9\sd{1.2} & 60.8\sd{1.0} & threshold (default) & 42.1\sd{0.6} & 60.3\sd{0.5} & 63.9\sd{1.2} & 60.8\sd{1.0} \\
$\theta = 0.3$ & 41.5\sd{0.5}\dt{-0.6} & 62.0\sd{0.4}\dtb{+1.7} & 60.6\sd{1.2}\dtb{-3.3} & 59.7\sd{1.0}\dt{-1.1} & top-1 (argmax) & 39.8\sd{0.3}\dtb{-2.3} & 53.6\sd{0.1}\dtb{-6.7} & 64.2\sd{1.2}\dt{+0.3} & 54.1\sd{0.9}\dtb{-6.7} \\
$\theta = 0.7$ & 41.7\sd{0.5}\dt{-0.4} & 58.2\sd{0.5}\dtb{-2.1} & 64.4\sd{1.1}\dt{+0.5} & 59.0\sd{1.4}\dt{-1.8} & top-2 & 30.9\sd{0.5}\dtb{-11.2} & 60.0\sd{0.3}\dt{-0.3} & 37.0\sd{0.8}\dtb{-26.9} & 37.5\sd{0.7}\dtb{-23.3} \\
$\theta = 0.9$ & 39.9\sd{0.3}\dtb{-2.2} & 54.1\sd{0.2}\dtb{-6.2} & 63.9\sd{0.9}\dt{0.0} & 53.7\sd{0.7}\dtb{-7.1} & top-3 & 25.4\sd{0.5}\dtb{-16.7} & 62.2\sd{0.4}\dtb{+1.9} & 27.2\sd{0.7}\dtb{-36.7} & 29.3\sd{0.6}\dtb{-31.5} \\
\bottomrule
\end{tabular}
}
\end{table}

%% file: tables/appendix/trainable_params.tex
\begin{table}[h]
\centering
\scriptsize
\tabcolsep=0.15cm
\setlength{\aboverulesep}{0pt}\setlength{\belowrulesep}{0pt}
\renewcommand{\arraystretch}{1.25}
\caption{\textbf{Trainable parameters and inference cost} of grounding models finetuned on \ourdataset{}.}
\label{tab:app-trainable}
\begin{tabular}{lccccc}
\toprule
\rowcolor{groupblue}
Method & Inference Params (B) & Trainable (B) & Trainable (\%) & Latency (s/img) & Peak VRAM (GB) \\
\midrule
GLaMM-7B~\citep{rasheed2024glamm} & 8.02 & 0.59 & 7.3 & 8.89 & 20.1 \\
Sa2VA-4B~\citep{yuan2025sa2va}     & 3.94  & 0.88  & 22.4 & 4.41 & 8.5 \\
Sa2VA-8B~\citep{yuan2025sa2va}     & 8.32  & 0.37  & 4.4  & 3.74 & 20.2 \\
SAMTok-4B~\citep{zhou2026samtok}   & 5.06  & 0.91   & 18.0 & 5.37 & 15.5 \\
SAMTok-8B~\citep{zhou2026samtok}   & 9.00  & 1.42 & 15.8 & 5.00 & 22.8 \\
\midrule
\rowcolor{rowgray}
\ourmodel{}-4B                     & 5.32  & 1.32 & 24.9 & 4.71 & 13.5 \\
\bottomrule
\end{tabular}
\end{table}